\documentclass[journal,onecolumn,12pt,draftclsnofoot]{IEEEtran}
\usepackage[margin=1in]{geometry}

\usepackage[utf8]{inputenc}
\usepackage{microtype}
\usepackage{graphicx}
\usepackage{booktabs} 
\usepackage{amsmath}
\usepackage{amsfonts}
\usepackage{amssymb}
\usepackage{amsthm}
\usepackage{color}
\usepackage{array}
\usepackage{cite}
\usepackage{enumerate}
\usepackage{graphicx}
\usepackage[hang]{subfigure}
\usepackage{epstopdf}
\usepackage{epsfig}
\usepackage{footmisc}
\usepackage{amsbsy}
\usepackage{bm}
\usepackage{bbm}
\usepackage{cases}
\usepackage{empheq}
\usepackage{multirow}
\usepackage{algorithmic}
\usepackage{comment}
\usepackage{times}
\usepackage{paralist}
\usepackage[ruled]{algorithm2e}
\usepackage{setspace}
\usepackage[breaklinks]{hyperref}
\usepackage{dsfont}

\newcommand{\note}[1]{[\textcolor{red}{\textit{#1}}]}

\newcommand{\bmx}{{\bm x}}

\newcommand{\bmh}{{\bm h}}
\newcommand{\bms}{{\bm s}}
\newcommand{\bmz}{{\bm z}}

\newcommand{\bmb}{{\bm b}}
\newcommand{\bmv}{{\bm v}}
\newcommand{\bmw}{{\bm w}}

\newcommand{\bmtheta}{\bm{\theta}}

\newcommand{\bmphi}{\bm{\phi}}

\newcommand{\bmsigma}{\bm{\sigma}}

\newcommand{\set}[1]{\ensuremath{\mathcal #1}}
\newcommand{\grad}[1]{\nabla #1}

\newcommand{\textH}{{\text{H}}}

\newcommand{\overra}{\overrightarrow}
\newcommand{\overla}{\overleftarrow}
\newcommand{\CNtoCP}{\mathsf{C}_{\text{N} \rightarrow \text{CP}}}
\newcommand{\CCPtoN}{\mathsf{C}_{\text{CP} \rightarrow \text{N}}}

\newcommand\GEMSNN{{GEM-SNN}}
\newcommand\MBSNN{{MB-SNN}}
\newcommand\IWSNN{{IW-SNN}}

\title{Multi-Sample Online Learning for \\ Probabilistic Spiking Neural Networks}

\author{
\IEEEauthorblockN{Hyeryung Jang and Osvaldo Simeone}
\thanks{H.~Jang and O.~Simeone are with King's Communications, Learning, and Information Processing (KCLIP) lab at the Centre for Telecommunications Research in the Department of Engineering, King's College London, London, United Kingdom (emails: hyeryung.jang@kcl.ac.uk, osvaldo.simeone@kcl.ac.uk).}
\thanks{This work was supported by the European Research Council (ERC) under the European Union's Horizon 2020 Research and Innovation programme (Grant Agreement No. 725731). It has also received suupport from Intel Labs as with the Intel's Neuromorphic Research Community (INRC) programme.}
}

\begin{document}

\bstctlcite{IEEEexample:BSTcontrol}

\maketitle

\begin{abstract}
Spiking Neural Networks (SNNs) capture some of the efficiency of biological brains for inference and learning via the dynamic, online, event-driven processing of binary time series. 
Most existing learning algorithms for SNNs are based on deterministic neuronal models, such as leaky integrate-and-fire, and rely on heuristic approximations of backpropagation through time that enforce constraints such as locality. 
In contrast, probabilistic SNN models can be trained directly via principled online, local, update rules that have proven to be particularly effective for resource-constrained systems. 
This paper investigates another advantage of probabilistic SNNs, namely their capacity to generate independent outputs when queried over the same input. 
It is shown that the multiple generated output samples can be used during inference to robustify decisions and to quantify uncertainty -- a feature that deterministic SNN models cannot provide. 
Furthermore, they can be leveraged for training in order to obtain more accurate statistical estimates of the log-loss training criterion, as well as of its gradient. 
Specifically, this paper introduces an online learning rule based on generalized expectation-maximization (GEM) that follows a three-factor form with global learning signals and is referred to as {\GEMSNN}. 
Experimental results on structured output memorization and classification on a standard neuromorphic data set demonstrate significant improvements in terms of log-likelihood, accuracy, and calibration when increasing the number of samples used for inference and training. 
\end{abstract}

\section{Introduction}
\label{sec:intro}

\subsection{Background}
\label{sec:background}

Much of the recent progress towards solving pattern recognition tasks in complex domains, such as natural images, audio, and text, has relied on parametric models based on Artificial Neural Networks (ANNs). 
It has been widely reported that ANNs often yields learning and inference algorithms with prohibitive energy and time requirements \cite{hao2019training, strubell2019energy}. 
This has motivated a renewed interest in exploring novel computational paradigms, such as Spiking Neural Networks (SNNs), that can capture some of the efficiency of biological brains for information encoding and processing. 
Neuromorphic computing platforms, including IBM's TrueNorth \cite{merolla2014million}, Intel's Loihi \cite{davies2018loihi}, and BrainChip's Akida \cite{brainchipakida}, implement SNNs instead of ANNs. 
Experimental evidence has confirmed the potential of SNNs implemented on neuromorphic chips in yielding energy consumption savings in many tasks of interest. 
For example, for a keyword spotting application, a Loihi-based implementation was reported to offer a $38.6 \times$ improvement in energy cost per inference as compared to state-of-the-art ANNs (see Fig.~$1$ in \cite{blouw2019benchmarking}). 
Other related experimental findings concern the identification of odorant samples under contaminants \cite{imam2020rapid} using SNNs via Loihi.

Inspired by biological brains, SNNs consist of neural units that process and communicate via sparse spiking signals over time, rather than via real numbers, over recurrent computing graphs \cite{neftci2019surrogate, jang2020reviewpt1, skatchkovsky2020reviewpt2, skatchkovsky2020reviewpt3}. Spiking neurons store and update a state variable, the membrane potential, that evolves over time as a function of past spike signals from pre-synaptic neurons. Most implementations are based on deterministic spiking neuron models that emit a spike when the membrane potential crosses a threshold. The design of training and inference algorithms for deterministic models need to address the non-differentiable threshold activation and the locality of the update permitted by neuromorphic chips \cite{neftci2019surrogate, davies2018loihi}. 
A training rule is said to a local if each neuron can update its synaptic weights based on locally available information during the causal operation of the network and based on scalar feedback signals. 
The non-differentiability problem is tackled by smoothing out the activation function \cite{huh2018gradient} or its derivative \cite{neftci2019surrogate}. In \cite{kaiser2020decolle, neftci2019surrogate, bohte2002error}, credit assignment is typically implemented via approximations of backpropagation through time (BPTT), such as random backpropagation, feedback alignment, or local randomized targets, that ensure the locality of the resulting update rule via {\em per-neuron scalar feedback signals}.

As an alternative, probabilistic spiking neural models based on the generalized linear model (GLM) \cite{nelder1972generalized} can be trained by directly maximizing a lower bound on the likelihood of the SNN producing desired outputs \cite{jimenez2014stochastic, brea2013matching, jang19:spm, zenke2018superspike}. 
The resulting learning algorithm implements an online learning rule that is local and only leverages global, rather than per-neuron, feedback signals. The rule leverages randomness for exploration and it was shown to be effective for resource-constrained systems \cite{jang20:vowel, skatchkovsky2020reviewpt2}.

\subsection{Main Contributions}

This paper investigates for the first time another potential advantage of probabilistic SNNs, namely their capability to generate multiple independent spiking signals when queried over the same input. 
This is clearly not possible with deterministic models which output functions of the input. 
We specifically consider two potential benefits arising from this property of probabilistic SNNs: {\em (i)} during {\em inference}, the multiple generated samples can be used to robustify the model's decisions and to quantify uncertainty; and {\em (ii)} during {\em training}, the independent samples can evaluate more accurate statistical estimates of the training criterion, such as the log-loss, as well as of its gradient, hence reducing the variance of the updates and improving convergence speed. 
Our main contributions are summarized as follows. 
\begin{itemize}[$\bullet$]
    \item During inference, we propose to apply independent runs of the SNNs for any given input in order to obtain multiple decisions. For the example of classification task, the decisions correspond to independent draws of the class index under the trained model. Using the multiple decisions allows more robust predictions to be made, and it enables a quantification of uncertainty. For classification, a majority rule can be implemented and uncertainty can be quantified via the histogram of the multiple decisions. 

    \item We introduce a multi-sample online learning rule that leverages generalized expectation-maximization (GEM) \cite{bishop2006pattern, simeone2018brief, tang2013sfnn} using multiple independent samples to approximate the log-loss bound via importance sampling. This yields a better statistical estimate of the log-loss and of its gradient as compared to conventional single-sample estimators \cite{tang2013sfnn}. The rule, referred to as {\GEMSNN}, follows a three-factor form with global per-sample learning signals computed by a central processor (CP) that communicates with all neurons (see Fig.~\ref{fig:model-learning}). We provide derivations of the rule and a study of communication loads between neurons and the CP. 
    
    \item We also derive alternative learning rules based on a mini-batch variant of the online learning rules in \cite{jimenez2014stochastic, brea2013matching, jang19:spm}, and on the principle of importance weighting \cite{burda2015importance, mnih2016variational,domke2018iwvi} in Appendix. 
    
    \item Finally, experimental results on structured output memorization and classification with a standard neuromorphic data set demonstrate the advantage of inference and learning rules for probabilistic SNNs that leverage multiple samples in terms of robust decision making, uncertainty quantification, log-likelihood, accuracy and calibration when increasing the number of samples used for inference and training. These benefits come at the cost of an increased communication load between neurons and the CP, as well as of the increased time and computational complexity required to generate the multiple samples.

\end{itemize}

\section{Probabilistic SNN Model}
\label{sec:model}

\begin{figure}[t!]
    \centering
    \includegraphics[height=0.22\columnwidth]{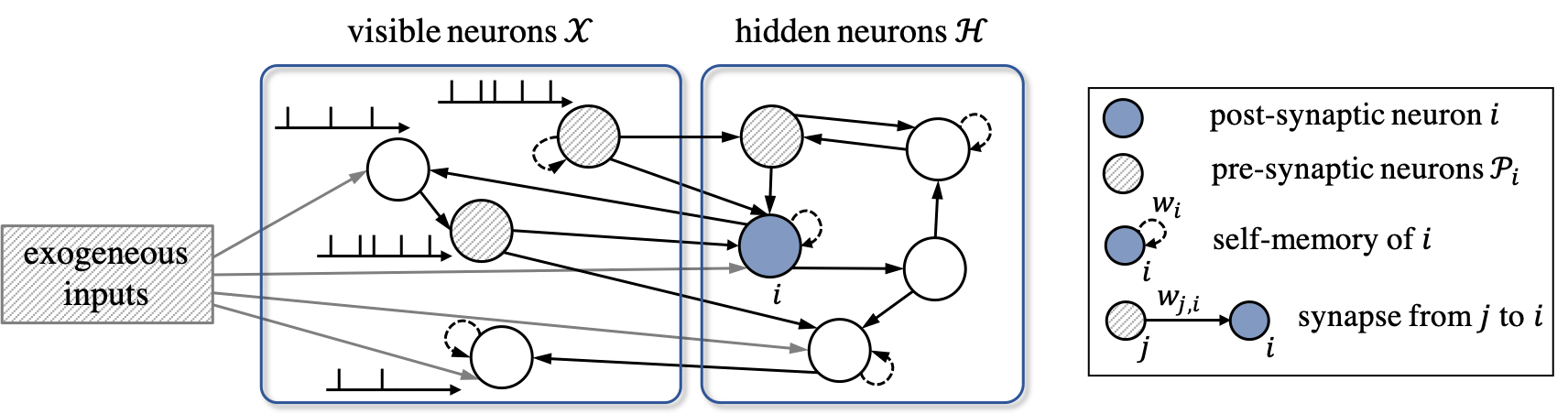}
    \caption{Architecture of an SNN with exogeneous inputs and $|\set{X}| = 4$ visible and $|\set{H}| = 5$ hidden spiking neurons -- the directed links between two neurons represent synaptic dependencies, while the self-loop links represent self-memory. The directed graph may have loops, indicating recurrent behavior.}
    \label{fig:model-topology}
    \vspace{-0.4cm}
\end{figure}

An SNN model is defined as a network connecting a set $\set{V}$ of spiking neurons via an arbitrary directed graph, which may have cycles. 
Following a discrete-time implementation of the probabilistic generalized linear neural model (GLM) for spiking neurons \cite{pillow2008spatio, jang19:spm}, at any time $t=1,2,\ldots$, each spiking neuron $i$ outputs a binary value $s_{i,t} \in \{0,1\}$, with ``$1$'' denoting the emission of a spike. We collect in vector $\bms_t = (s_{i,t}: i \in \set{V})$ the spiking signals emitted by all neurons $\set{V}$ at time $t$, and denote by $\bms_{\leq t} = (\bms_1, \ldots, \bms_t)$ the spike sequences of all neurons up to time $t$. As illustrated in Fig.~\ref{fig:model-topology}, each post-synaptic neuron $i$ receives past input spike signals $\bms_{\set{P}_i, \leq t-1}$ from the set $\set{P}_i$ of pre-synaptic neurons connected to it through directed links, known as synapses. With some abuse of notation, we include exogeneous inputs to a neuron $i$ (see Fig.~\ref{fig:model-topology}) in the set $\set{P}_i$ of pre-synaptic neurons.

\begin{figure}[t!]
    \centering
    \includegraphics[height=0.45\columnwidth]{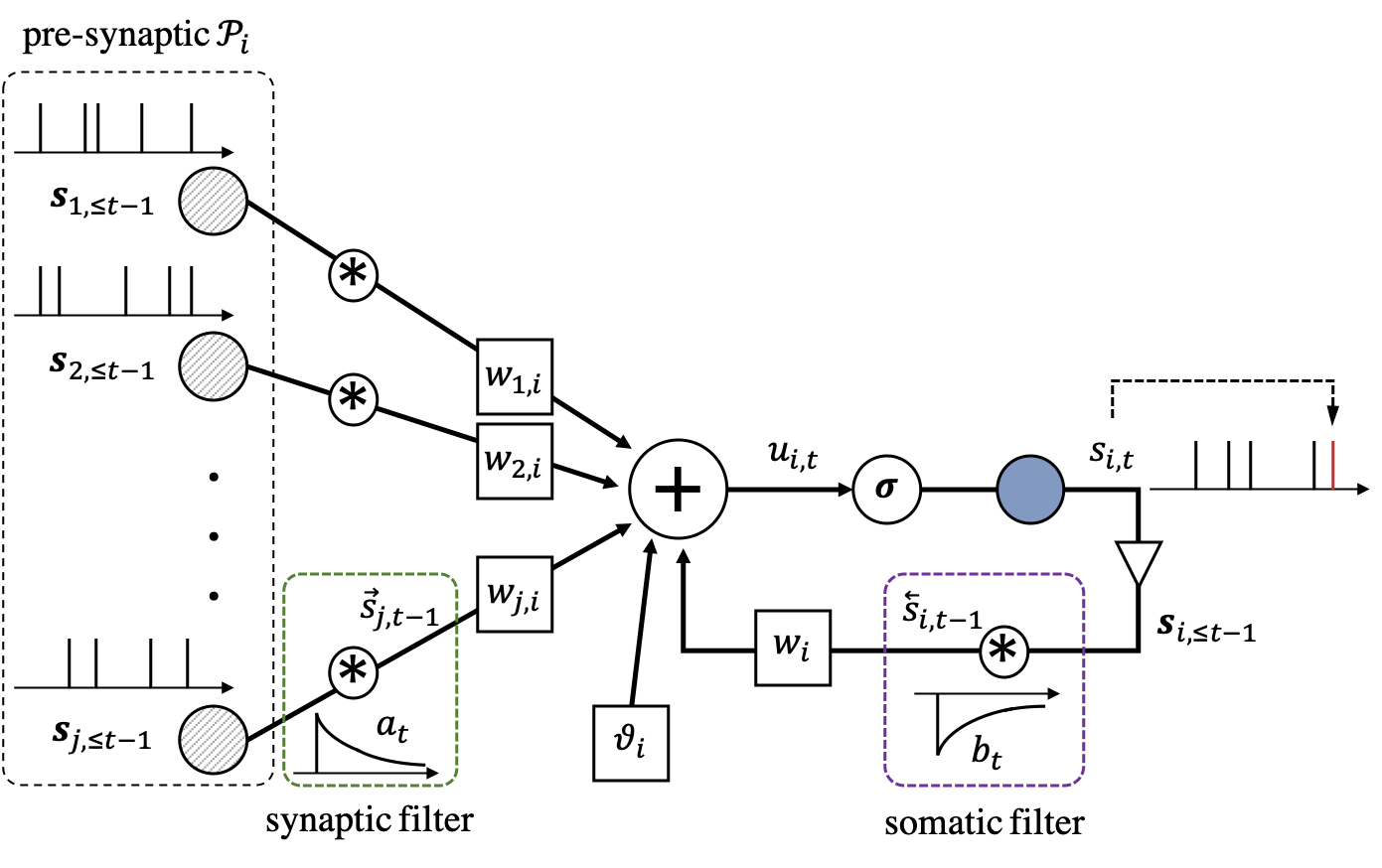}
    \vspace{-0.2cm}
    \caption{Illustration of the membrane potential $u_{i,t}$ model for a probabilistic SNN, with exponential synaptic and somatic filters. At each neuron $i$, the contribution of the synaptic trace from a pre-synaptic neuron $j \in \set{P}_i$ through synaptic filter $a_t$ are multiplied by the corresponding synaptic weights $w_{j,i}$, and the contribution of the somatic trace of a post-synaptic neuron $i$ through self-memory filter $b_t$ is multiplied by a weight $w_i$. The bias parameter $\vartheta_i$ is summed to obtain the membrane potential $u_{i,t}$, which is used to determine the spiking probability through the sigmoid function $\sigma(\cdot)$. }
    \label{fig:model-inference}
    \vspace{-0.5cm}
\end{figure}

As illustrated in Fig.~\ref{fig:model-inference}, the spiking probability of neuron $i$ at time $t$ conditioned on the value of the {\em membrane potential} $u_{i,t}$ is defined as
\begin{align} \label{eq:spike-prob-ind}
    p_{\bmtheta_i}(s_{i,t} = 1 | \bms_{\set{P}_i \cup i, \leq t-1}) = p_{\bmtheta_i}(s_{i,t} = 1 | u_{i,t}) = \sigma(u_{i,t}),
\end{align}
with $\sigma(x) = (1+\exp(-x))^{-1}$ being the sigmoid function. The membrane potential $u_{i,t}$ summarizes the effect of the past spike signals $\bms_{\set{P}_i,\leq t-1}$ from pre-synaptic neurons and of its past activity $\bms_{i, \leq t-1}$. From \eqref{eq:spike-prob-ind}, the negative log-probability of the output $s_{i,t}$ corresponds to the {\em binary cross-entropy loss} 
\begin{align} \label{eq:log-prob-ind}
    -\log p_{\bmtheta_i}(s_{i,t}| u_{i,t}) = \ell \big( s_{i,t}, \sigma(u_{i,t})\big) := -s_{i,t} \log \sigma(u_{i,t}) - (1-s_{i,t}) \log (1-\sigma(u_{i,t})).
\end{align}
The joint probability of the spike signals $\bms_{\leq T}$ up to time $T$ is defined using the chain rule as $p_{\bmtheta}(\bms_{\leq T}) = \prod_{t=1}^T \prod_{i \in \set{V}} p_{\bmtheta_i}(s_{i,t}|u_{i,t})$, where $\bmtheta = \{\bmtheta_i\}_{i \in \set{V}}$ is the model parameters, with $\bmtheta_i$ being the local model parameters of neuron $i$ as detailed below.

The membrane potential $u_{i,t}$ is obtained as the output of spatio-temporal synaptic filter, or kernel, $a_t$ and somatic filter $b_t$ \cite{doya2007bayesian, pillow2008spatio}. Specifically, each synapse $(j,i)$ from a pre-synaptic neuron $j \in \set{P}_i$ to a post-synaptic neuron $i$ computes the synaptic filtered trace 
\begin{align} \label{eq:synaptic-filtered-trace}
    \overra{s}_{j,t} = a_t \ast s_{j,t},
\end{align}
while the somatic filtered trace of neuron $i$ is computed as $\overla{s}_{i,t} = b_t \ast s_{i,t}$, where we denote by $f_t \ast g_t = \sum_{\delta > 0} f_{\delta} g_{t-\delta}$ the convolution operator. If the kernels are impulse responses of autoregressive filters, e.g., $\alpha$-functions with parameters, the filtered traces can be computed recursively without keeping track of windows of past spiking signals \cite{osogami15:DyBM}. It is also possible to assign multiple kernels to each synapse \cite{pillow2008spatio}. The membrane potential $u_{i,t}$ of neuron $i$ at time $t$ is finally given as the weighted sum
\begin{align} \label{eq:membrane-potential}
    u_{i,t} = \sum_{j \in \set{P}_i} w_{j,i} \overra{s}_{j,t-1} + w_i \overla{s}_{i,t-1} + \vartheta_i, 
\end{align}
where $w_{j,i}$ is a synaptic weight of the synapse $(j,i)$; $w_i$ is the self-memory weight; and $\vartheta_i$ is a bias, with $\bmtheta_i = \{ \{w_{j,i}\}_{j \in \set{P}_i}, w_i, \vartheta_i\}$ being the local model parameters of neuron $i$.

As illustrated in Fig.~\ref{fig:model-topology}, we divide the neurons of SNN into the disjoint subsets of visible $\set{X}$, and hidden, or latent, $\set{H}$ neurons, hence setting $\set{V} = \set{X} \cup \set{H}$. Using the notation above, we denote by $s_{i,t} = x_{i,t}$ for a visible neuron $i \in \set{X}$ and $s_{i,t} = h_{i,t}$ for a hidden neurons $i \in \set{H}$ the output at time $t$, as well as $\bms_t = (\bmx_t, \bmh_t)$ for the overall set of spike signals of all neurons $\set{V}$ at time $t$. 
The visible neurons represent the read-out layer that interfaces with end users or actuators. 
We will use the terms read-out layer and visible neurons interchangeably. 
The exogeneous input signals can either be recorded from neuromorphic sensors or they can instead be converted from a natural signal to a set of spiking signals by following different encoding rules, e.g., rate, time, or population encoding \cite{jang19:spm}. In a similar manner, output spiking signals of the visible neurons can either be fed directly to a neuromorphic actuator or they can be converted from spiking signals to natural signals by following pre-defined decoding principles, e.g., rate, time, or population decoding rules, in order to make the model's decision.

\section{Multi-Sample Inference}
\label{sec:multi-inference}

In this section, we describe ways in which multiple independent output samples produced by the SNN in response to a given input can be used to robustify decision making and quantify the uncertainty.

During inference, the SNN generally acts as a probabilistic sequence-to-sequence mapping between exogeneous inputs and outputs that is defined by the model parameters $\bmtheta$. 
While the output of the read-out layer for deterministic SNN models is a function of the exogeneous input, in probabilistic models, multiple applications of the SNN to the same input produce independent spiking outputs. 
Let us define as $K_I$ the number of independent spiking signals $\{\bmx_{\leq T}^k\}_{k=1}^{K_I}$ recorded at the read-out layer of visible neurons $\set{X}$ up to time $T$ for a given input. 
To elaborate on the use of the $K_I$ spiking signals $\{\bmx_{\leq T}^k\}_{k=1}^{K_I}$ for inference, we will focus in this section on classification tasks. 

Consider a classification task, in which each class $c \in \{1,\ldots,C\}$ is associated with a disjoint subset $\mathds{X}_c$ of output spiking signals at the read-out layer. 
Accordingly, the classifier chooses class $c$ if we have $\bmx_{\leq T} \in \mathds{X}_c$. 
For instance, with rate decoding, each class $c$ is assigned one neuron in the read-out layer of $C$ neurons, and the set $\mathds{X}_c$ contains all spiking signals $\bmx_{\leq T}$ such that 
\begin{align} \label{eq:rate-class}
    c = \arg \max_{c' \in \set{X}} \sum_{t=1}^T x_{c',t},
\end{align}
where $x_{c',t}$ is the output spike signal at time $t$ of the visible neuron $c' \in \set{X}$ corresponding to the class $c'$. 
This implies that the decoder chooses a class $c$ associated with the visible neuron producing the largest number of spikes.

The $K_I$ spiking signals $\{\bmx_{\leq T}^k\}_{k=1}^{K_I}$ of the visible neurons can be leveraged for classification by combining the $K_I$ decisions
\begin{align} \label{eq:multiple-class}
    \{\hat{c}^k: \bmx_{\leq T}^k \in \mathds{X}_{\hat{c}^k}\}_{k=1}^{K_I}.
\end{align}
In this example, one can adopt a majority decision rule whereby the final decision is obtained by selecting the class $\hat{c}$ that has received the most ``votes'', i.e., 
\begin{align} \label{eq:majority}
    \hat{c} = \arg \max_{c \in \set{X}} \sum_{k=1}^{K_I} \mathds{1}(c = \hat{c}^k),
\end{align}
with $\mathds{1}(E)$ being the indicator function of the event $E$. 
The final decision of majority rule \eqref{eq:majority} is generally more robust, being less sensitive to noise in the inference process. 

As an example, consider the problem of binary classification. If each decision fails with probability $P_e < 1/2$, the majority rule decodes erroneously with probability 
\begin{align} \label{eq:majority-error}
    P_{e,K_I} &= \sum_{k= \frac{K_I}{2} }^{K_I} \binom{K_I}{k} P_e^k (1-P_e)^{K_I-k} \leq \exp\Big( - 2 K_I \Big(\frac{1}{2} - P_e\Big)^2 \Big),
\end{align}
where the inequality follows by Hoeffding bound. It follows that the probability of error of the majority rule decreases exponentially with $K_I$. The same correlation applies with classification problems with any (finite) number of classes (see, e.g., \cite{moon2005error}). This will be demonstrated in Sec.~\ref{sec:experiments}.

Moreover, the $K_I$ spiking signals can be used to quantify the uncertainty of the decision. Focusing again on classification, for a given input, an ideal quantification of aleatoric uncertainty under the model requires computing the marginal distribution $p_{\bmtheta}(\bmx_{\leq T} \in \mathds{X}_c)$ for all classes $c \in \{1,\ldots,C\}$ (see, e.g., \cite{abdar2020uncertainty}). 
Probability $p_{\bmtheta}(\bmx_{\leq T} \in \mathds{X}_c)$ quantifies the level of confidence assigned by the model to each class $c$. 
Using the $K_I$ spiking signals $\{\bmx_{\leq T}^k \}_{k=1}^{K_I}$ produced by the visible neurons $\set{X}$ during $K_I$ independent runs, this probability can be estimated with the empirical average
\begin{align} \label{eq:majority-prob}
    p_{\bmtheta}(\bmx_{\leq T} \in \mathds{X}_c) \approx \frac{1}{K_I} \sum_{k=1}^{K_I} \mathds{1}(\bmx_{\leq T}^k \in \mathds{X}_c)
\end{align}
for each class $c$. This implies that each class is assigned a degree of confidence that depends on the fraction of the $K_I$ decisions that are made in its favour. 
We note that this approach is not feasible with deterministic models, which only provide individual decisions.

\section{Generalized EM Multi-Sample Online Learning for SNNs} 
\label{sec:multi-learning}

In this section, we discuss how the capacity of probabilistic SNNs to generate multiple independent samples at the output can be leveraged to improve the training performance in terms of convergence and accuracy. 
To this end, we first provide necessary background by reviewing the standard expectation-maximization (EM) algorithm and its Monte Carlo approximation known as generalized EM (GEM). 
These schemes are recalled for general probabilistic models. 
Next, we propose a novel learning rule that applies GEM to probabilistic SNNs. 
The novel rule, referred to as {\GEMSNN}, addresses the novel challenge of obtaining {\em online local} updates that can be implemented at each neuron as data is streamed through the SNN based on locally available information and limited global feedback.

As in \cite{jimenez2014stochastic, brea2013matching, jang19:spm}, we aim at maximizing the likelihood of obtaining a set of desired output spiking signals at the read-out layer. 
The desired output sequences are specified by the training data as spiking sequences $\bmx_{\leq T}$ of some duration $T > 0$ in response to given exogeneous inputs. In contrast, the spiking behavior of the hidden neurons is left unspecified by the data, and it should be adapted to ensure the desired outputs of the visible neurons.

Mathematically, we focus on the maximum likelihood (ML) problem
\begin{align} \label{eq:ml}
    \min_{\bmtheta}~ - \log p_{\bmtheta}(\bmx_{\leq T}),
\end{align}
where $\bmx_{\leq T}$ is a sequence of desired outputs for the visible neurons. 
The likelihood of the data, $p_{\bmtheta}(\bmx_{\leq T})$, is obtained via marginalization over the hidden spiking signals as 
\begin{align} \label{eq:ml-marginal}
    \log p_{\bmtheta}(\bmx_{\leq T}) = \log \sum_{\bmh_{\leq T}} p_{\bmtheta}(\bmx_{\leq T}, \bmh_{\leq T}).
\end{align} 
This marginalization requires summing over the $2^{|\set{H}|T}$ possible values of the hidden neurons, which is practically infeasible. 
As we will discuss, we propose a multi-sample online learning rule that estimates the expectation over the hidden neurons' outputs by using $K$ independent realizations of the outputs $\{\bmh_{\leq T}^k\}_{k=1}^K$ of the hidden neurons, which are obtained via $K$ independent forward passes through the SNN for a given input.

\subsection{Expectation-Maximization Algorithm} \label{sec:EM}

We start by recalling the operation of the standard EM algorithm \cite{dempster1977maximum} for general probabilistic models of the form $p_{\bmtheta}(\bmx,\bmh)$, where $\bmh$ is a discrete latent random vector. 
In a manner analogous to \eqref{eq:ml}-\eqref{eq:ml-marginal}, in general probabilistic models, one aims at minimizing the log-loss, i.e., $\min_{\bmtheta} \big(- \log p_{\bmtheta}(\bmx)\big)$, where the log-likelihood of the data is obtained via marginalization over the hidden vector $\bmh$ as $\log p_{\bmtheta}(\bmx) = \log \sum_{\bmh} p_{\bmtheta}(\bmx,\bmh)$. The EM algorithm introduces an auxiliary distribution $q(\bmh)$, known as {\em variational posterior}, and focuses on the minimization of the upper bound 
\begin{align} \label{eq:general-elbo}
    -\log p_{\bmtheta}(\bmx) 
    &= -\sum_{\bmh} p_{\bmtheta}(\bmh|\bmx) \log \frac{p_{\bmtheta}(\bmx,\bmh)}{p_{\bmtheta}(\bmh|\bmx)} \cr 
    &\leq - \sum_{\bmh} q(\bmh) \log \frac{p_{\bmtheta}(\bmx,\bmh)}{q(\bmh)} = -\mathbb{E}_{q(\bmh)}\Big[ \log p_{\bmtheta}(\bmx,\bmh) \Big] - \text{H}(q(\bmh)),
\end{align}
where $\text{H}(q(\bmh))$ is the entropy of the distribution $q(\bmh)$. The upper bound \eqref{eq:general-elbo} is known as {\em variational free energy} (or negative evidence lower bound, ELBO), and is tight when the auxiliary distribution $q(\bmh)$ equals the exact posterior $p_{\bmtheta}(\bmh|\bmx)$.

Based on this observation, the EM algorithm iterates between optimizing over $q(\bmh)$ and over $\bmtheta$ in separate phases known as E and M steps. 
To elaborate, for a fixed current iterate $\bmtheta_\text{old}$ of the model parameters, the E step minimizes the bound \eqref{eq:general-elbo} over $q(\bmh)$, obtaining $q(\bmh) = p_{\bmtheta_\text{old}}(\bmh|\bmx)$. Then, it computes the expected log-loss
\begin{subequations} \label{eq:EM}
\begin{align} \label{eq:EM-estep}
    \set{L}_{\bmx}(\bmtheta,\bmtheta_\text{old}) := -\mathbb{E}_{p_{\bmtheta_\text{old}}(\bmh|\bmx)}\Big[ \log p_{\bmtheta}(\bmx,\bmh)\Big].
\end{align}
The M step finds the next iterate for parameters $\bmtheta$ by minimizing the bound \eqref{eq:general-elbo} for fixed $q(\bmh) = p_{\bmtheta_\text{old}}(\bmh|\bmx)$, which corresponds to the problem 
\begin{align} \label{eq:EM-mstep}
    \bmtheta_\text{new} = \arg \min_{\bmtheta}~ \set{L}_{\bmx}(\bmtheta,\bmtheta_\text{old}).
\end{align}
\end{subequations}

The implementation of the EM algorithm \eqref{eq:EM} is computationally demanding for problems of practical size. In fact, the E step in \eqref{eq:EM-estep} entails the computation of the posterior $p_{\bmtheta_\text{old}}(\bmh|\bmx)$ of the hidden vector at the current iterate $\bmtheta_\text{old}$; and the M step requires solving the stochastic optimization problem \eqref{eq:EM-mstep}.

\subsection{Generalized Expectation-Maximization Algorithm} \label{sec:GEM}

We now review a Monte Carlo approximation of the EM algorithm, known as GEM \cite{neal1998view, tang2013sfnn} that tackles the two computational issues identified above. 
First, the expected log-loss $\set{L}_{\bmx}(\bmtheta,\bmtheta_\text{old})$ in \eqref{eq:EM-estep} is estimated via importance sampling by using $K$ independent samples $\{\bmh^k\}_{k=1}^K$ of the hidden variables drawn from the current prior distribution, i.e., $\bmh^k \sim p_{\bmtheta_\text{old}}(\bmh)$ for $k=1,\ldots,K$. Specifically, GEM considers the unbiased estimate 
\begin{align} \label{eq:gem-estep}
    \set{L}_{\bmx}(\bmtheta,\bmtheta_\text{old}) &= - \mathbb{E}_{p_{\bmtheta_\text{old}}(\bmh)} \bigg[ \frac{p_{\bmtheta_\text{old}}(\bmh|\bmx)}{p_{\bmtheta_\text{old}}(\bmh)} \log p_{\bmtheta}(\bmx,\bmh) \bigg] \cr
    &\approx -\sum_{k=1}^K  w^k \cdot \log p_{\bmtheta}(\bmx,\bmh^k) := \set{L}^K_{\bmx}(\bmtheta,\bmtheta_\text{old}),
\end{align}
where the {\em importance weight} $w^k$ of the sample $\bmh^k$ is defined as
\begin{align} \label{eq:importance-weight-gem}
    w^k:= \frac{1}{K} \cdot \frac{p_{\bmtheta_\text{old}}(\bmh^k|\bmx)}{p_{\bmtheta_\text{old}}(\bmh^k)} = \frac{1}{K} \cdot \frac{p_{\bmtheta_\text{old}}(\bmx|\bmh^k)}{p_{\bmtheta_\text{old}}(\bmx)}. 
\end{align}
The marginal $p_{\bmtheta_\text{old}}(\bmx)$ in \eqref{eq:importance-weight-gem} also requires averaging over the hidden variables and is further approximated using the same samples $\{\bmh^k\}_{k=1}^K$ as
\begin{align}
    p_{\bmtheta_\text{old}}(\bmx) \approx \frac{1}{K} \sum_{k'=1}^K p_{\bmtheta_\text{old}}(\bmx|\bmh^{k'}).
\end{align}
Accordingly, the weights in \eqref{eq:importance-weight-gem} are finally written as
\begin{align} \label{eq:importance-weight-gem-sm}
    w^k = \frac{p_{\bmtheta_\text{old}}(\bmx|\bmh^k)}{\sum_{k'=1}^K p_{\bmtheta_\text{old}}(\bmx|\bmh^{k'})} =  \bmsigma_\text{SM}^k\big( \bmv \big).
\end{align}
In \eqref{eq:importance-weight-gem-sm}, we have defined the SoftMax function 
\begin{align} \label{eq:softmax}
    \bmsigma_\text{SM}^k\big( \bmv \big) = \frac{\exp \big( v^k \big)}{\sum_{k'=1}^K \exp \big( v^{k'} \big) }, ~\text{for}~ k=1,\ldots,K
\end{align}
and denoted as $v^k:= \log p_{\bmtheta_\text{old}}(\bmx|\bmh^k)$ the log-probability of producing the observation $\bmx$ given the $k$th realization $\bmh^k$ of the hidden vector. We have also introduced in vector $\bmv = (v^1,\ldots,v^K)$. Note that the importance weights sum to $1$ by definition. 
The variance of the estimate given by \eqref{eq:gem-estep} and \eqref{eq:importance-weight-gem-sm} generally depends on how well the current latent prior $p_{\bmtheta_\text{old}}(\bmh)$ represents the posterior $p_{\bmtheta_\text{old}}(\bmh|\bmx)$ \cite{neal1998view, tang2013sfnn}.

In the M step, GEM relies on numerical optimization of $\set{L}_{\bmx}^K(\bmtheta,\bmtheta_\text{old})$ in \eqref{eq:gem-estep} with \eqref{eq:importance-weight-gem-sm} via gradient descent yielding the update 
\begin{align} \label{eq:gem-mstep}
    \bmtheta &\leftarrow \bmtheta - \eta \cdot \grad_{\bmtheta} \set{L}^K_{\bmx}(\bmtheta,\bmtheta_\text{old}) \big|_{\bmtheta = \bmtheta_\text{old}} \cr 
    &= \bmtheta + \eta \cdot \sum_{k=1}^K \bmsigma_\text{SM}^k(\bmv) \grad_{\bmtheta} \log p_{\bmtheta}(\bmx,\bmh^k) \big|_{\bmtheta = \bmtheta_\text{old}}, \quad
\end{align}
where $\eta > 0$ is the learning rate. From \eqref{eq:gem-mstep}, the gradient is a weighted sum of the $K$ partial derivatives $\{\grad_{\bmtheta} \log p_{\bmtheta}(\bmx,\bmh^k)\}_{k=1}^K$, with the importance weight $\bmsigma_\text{SM}^k(\bmv)$ measuring the relative effectiveness of the $k$th realization $\bmh^k$ of the hidden variables in generating the observation $\bmx$.

\subsection{Derivation of {\GEMSNN}} \label{sec:GEM-SNN}

In this section, we propose an online learning rule for SNNs that leverages $K$ independent samples from the hidden neurons by adapting the GEM algorithm introduced in Sec.~\ref{sec:GEM} to probabilistic SNNs. 
To start, we note that a direct application of GEM to the learning problem \eqref{eq:ml} for SNNs is undesirable for two reasons. First, the use of the prior $p_{\bmtheta_\text{old}}(\bmh_{\leq T})$ as proposal distribution in \eqref{eq:gem-estep} is bound to cause the variance of the estimate \eqref{eq:gem-estep} to be large. This is because the prior distribution $p_{\bmtheta_\text{old}}(\bmh_{\leq T})$ is likely to be significantly different from the posterior $p_{\bmtheta_\text{old}}(\bmh_{\leq T}|\bmx_{\leq T})$, due to the temporal dependence of the outputs of visible neurons and hidden neurons. 
Second, a direct application of GEM would yield a batch update rule, rather than the desired online update. We now tackle these two problems in turn.

{\em 1) Sampling distribution.} To address the first issue, following \cite{jimenez2014stochastic, brea2013matching, jang19:spm}, we propose to use as proposal distribution the {\em causally conditioned distribution} $p_{\bmtheta_\text{old}}(\bmh_{\leq T}||\bmx_{\leq T})$, where we have used the notation \cite{kramer1998directed}
\begin{align} \label{eq:causally-cond}
    p_{\bmtheta^\text{H}}(\bmh_{\leq T} || \bmx_{\leq T}) &= \prod_{t=1}^T p_{\bmtheta}(\bmh_{t} | \bmx_{\leq t}, \bmh_{\leq t-1}) = \prod_{t=1}^T \prod_{i \in \set{H}} p_{\bmtheta_i}(h_{i,t} | u_{i,t}).
\end{align}
Furthermore, distribution \eqref{eq:causally-cond} is easy to sample from, since the samples from \eqref{eq:causally-cond} can be directly obtained by running the SNN in standard forward mode. 
In contrast to true posterior distribution $p_{\bmtheta}(\bmh_{\leq T}|\bmx_{\leq T}) = \prod_{t=1}^T p_{\bmtheta}(\bmh_t |\bmx_{\leq T},\bmh_{\leq t-1})$, the causally conditioned distribution \eqref{eq:causally-cond} ignores the stochastic dependence of the hidden neurons' signals $\bmh_t$ at time $t$ on the future values of the visible neurons' signals $\bmx_{> t}$. Nevertheless, it does capture the dependence of $\bmh_t$ on past signals $\bmx_{\leq t}$, and is therefore generally a much better approximation of the posterior as compared to the prior.

With this choice as the proposal distribution, during training, we obtain the $K$ independent realizations of the outputs $\bmh_{\leq T}^{1:K} = \{\bmh_{\leq T}^k\}_{k=1}^K$ of the hidden neurons via $K$ independent forward passes through the SNN at the current parameters $\bmtheta_\text{old}$. 
Denoting by $\bmtheta^\text{X} = \{\bmtheta_i\}_{i \in \set{X}}$ and $\bmtheta^\text{H} = \{\bmtheta_i\}_{i \in \set{H}}$ the collection of model parameters for visible $\set{X}$ and hidden $\set{H}$ neurons, respectively, we note that we have the decomposition
\begin{align}
    p_{\bmtheta}(\bmx_{\leq T},\bmh_{\leq T}) = p_{\bmtheta^\text{H}}(\bmh_{\leq T}||\bmx_{\leq T})p_{\bmtheta^\text{X}}(\bmx_{\leq T}||\bmh_{\leq T-1}),   
\end{align}
with $p_{\bmtheta^\text{X}}(\bmx_{\leq T}||\bmh_{\leq T-1}) = \prod_{t=1}^T \prod_{i \in \set{X}} p_{\bmtheta_i}(x_{i,t}|u_{i,t})$ \cite{kramer1998directed}. 
By leveraging the $K$ samples $\bmh_{\leq T}^{1:K}$ via importance sampling as in \eqref{eq:gem-estep}-\eqref{eq:importance-weight-gem}, we accordingly compute the approximate importance weight \eqref{eq:importance-weight-gem-sm} as 
\begin{align} \label{eq:posterior-approx}
    \frac{1}{K} \cdot \frac{ p_{\bmtheta_\text{old}}(\bmh_{\leq T}^k | \bmx_{\leq T}) }{ p_{\bmtheta_\text{old}^\text{H}}(\bmh_{\leq T}^k || \bmx_{\leq T})} &= \frac{1}{K} \cdot \frac{p_{\bmtheta_\text{old}}(\bmx_{\leq T}, \bmh_{\leq T}^k)}{p_{\bmtheta_\text{old}}(\bmx_{\leq T}) \cdot p_{\bmtheta_\text{old}^\text{H}}(\bmh_{\leq T}^k || \bmx_{\leq T})} \cr
    &\approx \frac{ p_{\bmtheta_\text{old}^\text{X}} (\bmx_{\leq T} || \bmh_{\leq T-1}^k)}{ \sum_{k'=1}^K p_{\bmtheta_\text{old}^\text{X}}(\bmx_{\leq T} || \bmh_{\leq T-1}^{k'}) } 
    = \bmsigma_\text{SM}^k\Big( \bmv_{\bmtheta_\text{old}^\text{X},T} \Big). \qquad
\end{align}
In \eqref{eq:posterior-approx}, $v_{\bmtheta_\text{old}^\text{X},T}^k$ represents the log-probability of producing the visible neurons' target signals $\bmx_{\leq T}$ causally conditioned on $\bmh_{\leq T-1}^k$, i.e., 
\begin{align} \label{eq:gem-log-weight-ls}
    v_{\bmtheta_\text{old}^\text{X},T}^k &:= \log p_{\bmtheta_\text{old}^\text{X}}(\bmx_{\leq T}||\bmh_{\leq T-1}^k) = -\sum_{t=1}^T \sum_{i \in \set{X}} \ell\big( x_{i,t},\sigma(u_{i,t}^k)\big).
\end{align}
These probabilities are collected in vector $\bmv_{\bmtheta_\text{old}^\text{X},T} = \big(v_{\bmtheta_\text{old}^\text{X},T}^1, \ldots, v_{\bmtheta_\text{old}^\text{X},T}^K\big)$. 
As a result, we have the MC estimate $\set{L}_{\bmx_{\leq T}}^K(\bmtheta,\bmtheta_\text{old})$ of the upper bound 
\begin{align} \label{eq:gem-snn-elbo-T}
    &\set{L}_{\bmx_{\leq T}}^K(\bmtheta,\bmtheta_\text{old}) = \sum_{k=1}^K \bigg( \bmsigma_\text{SM}^k \Big( \bmv_{\bmtheta_\text{old}^\text{X},T} \Big) \cdot \sum_{t=1}^T \sum_{i \in \set{V}} \ell\big( s_{i,t}^k, \sigma(u_{i,t}^k)\big) \bigg). 
\end{align}

{\em 2) Online learning rule.} 
We now consider the second problem, namely that of obtaining an online rule from the minimization of \eqref{eq:gem-snn-elbo-T}. In online learning, the model parameters $\bmtheta$ are updated at each time $t$ based on the data $\bmx_{\leq t}$ observed so far. 
To this end, we propose to minimize at each time $t$ the discounted version of the upper bound $\set{L}_{\bmx_{\leq T}}^K(\bmtheta,\bmtheta_\text{old})$ in \eqref{eq:gem-snn-elbo-T} given as 
\begin{align}
\label{eq:gem-snn-elbo}
    \set{L}^{K,\gamma}_{\bmx_{\leq t}}(\bmtheta, \bmtheta_\text{old}) &:= \sum_{k=1}^K \bigg( \bmsigma_\text{SM}^k\Big( \bmv_{\bmtheta_\text{old}^\text{X},t}^{\gamma} \Big) \cdot \sum_{t'=0}^{t-1} \gamma^{t'} \sum_{i \in \set{V}} \ell\big( s_{i,t-t'}^k, \sigma(u_{i,t-t'}^k)\big) \bigg), 
\end{align}
where $\gamma \in (0,1)$ is a discount factor. Furthermore, we similarly compute a discounted version of the log-probability $v_{\bmtheta_\text{old}^\text{X},t}^k$ at time $t$ by using temporal average operator as
\begin{align} \label{eq:gem-log-weight}
    v_{\bmtheta_\text{old}^\text{X},t}^{k,\gamma} &:= -\sum_{t'=0}^{t-1} \gamma^{t'} \sum_{i \in \set{X}} \ell\big( x_{i,t-t'},\sigma(u_{i,t-t'}^k)\big) =
    \Big\langle -\sum_{i \in \set{X}} \ell\big( x_{i,t},\sigma(u_{i,t}^k)\big) \Big\rangle_\gamma.
\end{align}
In \eqref{eq:gem-log-weight}, we denoted as $\langle f_t\rangle_\kappa$ the temporal average operator of a time sequence $\{f_t\}_{t \geq 1}$ with some constant $\kappa \in (0,1)$ as $\langle f_t\rangle_\kappa = \kappa \cdot \langle f_{t-1}\rangle_\kappa + f_t$ with $\langle f_0\rangle_\kappa = 0$. 
In a similar way, the batch processing for computing the log-loss in \eqref{eq:gem-snn-elbo} can be obtained by $\big\langle \sum_{i \in \set{V}} \ell\big( s_{i,t}^k, \sigma(u_{i,t}^k)\big) \big\rangle_\gamma$, with $s_{i,t}^k = x_{i,t}$ for visible neuron $i \in \set{X}$ and $s_{i,t}^k = h_{i,t}^k$ for hidden neuron $i \in \set{H}$. 
We note that $\set{L}_{\bmx_{\leq T}}^{K,\gamma}(\bmtheta,\bmtheta_\text{old}) \rightarrow \set{L}_{\bmx_{\leq T}}^K(\bmtheta,\bmtheta_\text{old})$ as $\gamma \rightarrow 1$.

The minimization of the bound $\set{L}_{\bmx_{\leq t}}^{K,\gamma}(\bmtheta,\bmtheta_\text{old})$ in \eqref{eq:gem-snn-elbo} via stochastic gradient descent (SGD) corresponds to the generalized M step. 
The resulting online learning rule, which we refer to {\GEMSNN}, updates the parameters $\bmtheta$ in the direction of the gradient $-\grad_{\bmtheta} \set{L}_{\bmx_{\leq t}}^{K,\gamma}(\bmtheta,\bmtheta_\text{old})$, yielding the update rule at time $t$ 
\begin{align} \label{eq:gem-snn-update-simple}
    \bmtheta \leftarrow \bmtheta - \eta \cdot \sum_{k=1}^K & \bmsigma_\text{SM}^k\Big( \bmv_{\bmtheta_\text{old}^\text{X},t}^\gamma \Big) \cdot \Big\langle \sum_{i \in \set{V}} \grad_{\bmtheta} \ell\big( s_{i,t}^k, \sigma(u_{i,t}^k)\big) \Big|_{\bmtheta = \bmtheta_\text{old}} \Big\rangle_\gamma, 
\end{align}
with a learning rate $\eta$.

\subsection{Summary of {\GEMSNN}}

\begin{figure}[t!]
    \centering
    \includegraphics[height=0.43\columnwidth]{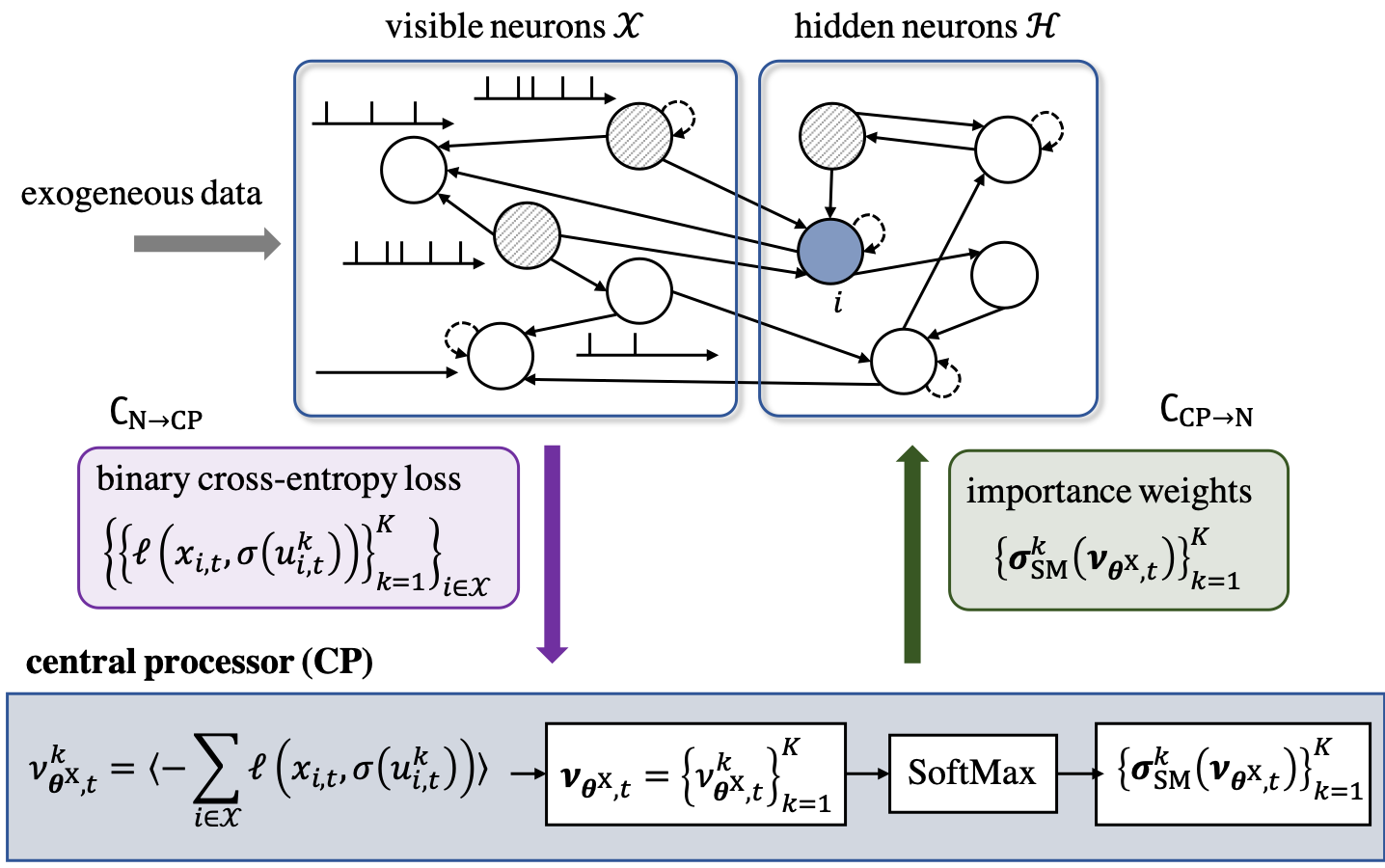}
    \vspace{-0.3cm}
    \caption{Illustration of the proposed multi-sample online learning scheme for SNNs. In {\GEMSNN}, learnable model parameters $\bmtheta$ are adapted based on data fed into the visible neurons $\set{X}$ with the help of the stochastic spiking behavior of hidden neurons $\set{H}$. A central processor (CP) collects information from all $K$ samples of the visible neurons $\set{X}$, with entailing a communication load $\CNtoCP = K|\set{X}|$; computes importance weights of samples; and sends them to all neurons, with entailing a broadcast communication load $\CCPtoN = K(|\set{X}|+|\set{H}|)$, in order to guide the update of model parameters $\bmtheta$. }
    \label{fig:model-learning}
    \vspace{-0.3cm}
\end{figure}

As illustrated in Fig.~\ref{fig:model-learning}, the rule {\GEMSNN} can be summarized as follows. At each time $t=1,2,\ldots,$ a central processor (CP) collects the binary cross-entropy values $\{\{\ell\big( x_{i,t}, \sigma(u_{i,t}^k)\big)\}_{k=1}^K\}_{i \in \set{X}}$ from all visible neurons $i \in \set{X}$ from $K$ parallel independent runs of the SNN. 
It then computes the log-probabilities of the samples $\bmv_{\bmtheta^\text{X},t}^\gamma$ with a constant $\gamma \in (0,1)$ as in \eqref{eq:gem-log-weight}, as well as the corresponding importance weights of the samples using the SoftMax function $\bmsigma_\text{SM}(\cdot)$ in \eqref{eq:softmax}. Then, the SoftMax values $\{\bmsigma_\text{SM}^k\big( \bmv_{\bmtheta^\text{X},t}^\gamma \big)\}_{k=1}^K$ are fed back from the CP to all neurons. Finally, each neuron $i \in \set{V}$ updates the local model parameters $\bmtheta_i$ as $\bmtheta_i \leftarrow \bmtheta_i + \eta \cdot \Delta \bmtheta_i$, where each component of $\Delta \bmtheta_i$ is given as 
\begin{align} \label{eq:gem-update}
    \Delta w_{j,i} &= \sum_{k=1}^K \bmsigma_\text{SM}^k\Big( \bmv_{\bmtheta^\text{X},t}^\gamma \Big) \cdot \Big\langle \big( s_{i,t}^k - \sigma(u_{i,t}^k) \big) \cdot \overra{s}_{j,t-1}^{k} \Big\rangle_\gamma, 
    \cr
    \Delta w_i &= \sum_{k=1}^K \bmsigma_\text{SM}^k\Big( \bmv_{\bmtheta^\text{X},t}^\gamma \Big) \cdot \Big\langle \big( s_{i,t}^k - \sigma(u_{i,t}^k) \big) \cdot \overla{s}_{i,t-1}^k \Big\rangle_\gamma, 
    \cr
    \Delta \vartheta_i &= \sum_{k=1}^K \bmsigma_\text{SM}^k\Big( \bmv_{\bmtheta^\text{X},t}^\gamma \Big) \cdot \Big\langle s_{i,t}^k - \sigma(u_{i,t}^k) \Big\rangle_\gamma, 
\end{align}
where we have used standard expressions for the derivatives of the cross-entropy loss \cite{jang19:spm,jang19:def}, and we set $s_{i,t}^k = x_{i,t}$ for visible neuron $i \in \set{X}$ and $s_{i,t}^k = h_{i,t}^k$ for hidden neuron $i \in \set{H}$. As we will explain next, the update \eqref{eq:gem-update} requires only local information at each neuron, except for the $K$ SoftMax values broadcast by the CP. The overall algorithm is detailed in Algorithm~\ref{alg:gem-snn}.

\DecMargin{2em}
\begin{algorithm}[t]
\caption{{\GEMSNN}}
\label{alg:gem-snn}
\begin{algorithmic}[1]
   \STATE {\bfseries Input:} 
   Training data $\bmx_{\leq t}$, number of samples for training $K$, discount factor $\gamma$, and learning rate $\eta$
   
   \STATE{\bfseries Output:} learned model parameters $\bmtheta$
   
   \vspace{-0.2cm}
   \hrulefill
   
   \STATE {\bf initialize} parameters $\bmtheta$
    
    \FOR{each time $t=1,2,\ldots$}
    
    \smallskip
    \STATE \texttt{$K$ independent forward passes:} 
    
    \smallskip
    \FOR{$k=1,\ldots,K$}
    \STATE each neuron $i \in \set{V}$ computes the filtered traces $\{\overra{s}_{i,t-1}^k, \overla{s}_{i,t-1}^k\}$ and computes the membrane potential $u_{i,t}^k$ from \eqref{eq:membrane-potential}
    
    
    \smallskip
    \STATE each hidden neuron $i \in \set{H}$ emits a spike $h_{i,t}^k = 1$ with probability $\sigma(u_{i,t}^k)$ as in \eqref{eq:spike-prob-ind}

    \ENDFOR
    
    \smallskip
    \STATE a central processor (CP) collects the binary cross-entropy values $\big\{\ell\big(x_{i,t},\sigma(u_{i,t}^k)\big)\big\}_{k=1}^K$ in \eqref{eq:log-prob-ind} from all visible neurons $i \in \set{X}$, and computes a discounted version of the log-probability $\bmv_{\bmtheta^\text{X},t}^\gamma$ from \eqref{eq:gem-log-weight}
    
    \smallskip
    \STATE the CP computes the approximate importance weights $\big\{\bmsigma_\text{SM}^k\big( \bmv_{\bmtheta^\text{X},t}^\gamma \big)\big\}_{k=1}^K$ using the SoftMax function, and feeds back them to all neurons $\set{V}$
    
    \smallskip
    \STATE \texttt{parameter update:}
    \STATE each neuron $i \in \set{V}$ updates the local model parameters $\bmtheta_i$ as
    \begin{align*}
        \bmtheta_i \leftarrow \bmtheta_i + \eta \cdot \sum_{k=1}^K \bmsigma_\text{SM}^k\Big( \bmv_{\bmtheta^\text{X},t}^\gamma\Big) \cdot \grad_{\bmtheta_i} \ell\big( s_{i,t}^k, \sigma(u_{i,t}^k)\big),
    \end{align*}
    with $s_{i,t}^k = x_{i,t}$ for $i \in \set{X}$ and $s_{i,t}^k = h_{i,t}^k$ for $i \in \set{H}$
    \ENDFOR
\end{algorithmic}
\end{algorithm}
\IncMargin{2em}

\subsection{Interpreting {\GEMSNN}}

The update rule {\GEMSNN} \eqref{eq:gem-update} follows a standard three-factor format implemented by most neuromorphic hardware \cite{davies2018loihi, fremaux2016neuromodulated}: A synaptic weight $w_{j,i}$ from pre-synaptic neuron $j$ to a post-synaptic neuron $i$ is updated as  
\begin{align} \label{eq:three-factor}
    w_{j,i} \leftarrow w_{j,i} + \eta \cdot \sum_{k=1}^K \textsf{learning signal}^k \cdot \big\langle \textsf{pre}_j^k \cdot \textsf{post}_i^k \big\rangle,
\end{align}
where $\eta$ is the learning rate. The three-factor rule \eqref{eq:three-factor} sums the contribution from the $K$ samples, with each contribution depending on three factors: $\textsf{pre}_j^k$ and $\textsf{post}_i^k$ represents the activity of the pre-synaptic neuron $j$ and of the post-synaptic neuron $i$, respectively; and $\textsf{learning signal}^k$ determines the sign and magnitude of the contribution of the $k$th sample to the learning update. The update rule \eqref{eq:three-factor} is hence local, with the exception of the global learning signals.

In particular, in \eqref{eq:gem-update}, for each neuron $i \in \set{V}$, the gradients $\grad_{\bmtheta_i} \set{L}_{\bmx_{\leq t}}^{K,\gamma}(\cdot)$ contain the synaptic filtered trace $\overra{s}_{j,t}^k$ of {\em pre-synaptic} neuron $j$; the somatic filtered trace $\overla{s}_{i,t}^k$ of {\em post-synaptic} neuron $i$ and the {\em post-synaptic} error $s_{i,t}^k - \sigma(u_{i,t}^k)$; and the {\em global learning signals} $\big\{\bmsigma_\text{SM}^k\big(\bmv_{\bmtheta^\text{X},t}^\gamma \big)\big\}_{k=1}^K$. 
Accordingly, the importance weights computed using the SoftMax function can be interpreted as the common learning signals for all neurons, with the contribution of each $k$th sample being weighted by $\bmsigma_\text{SM}^k\big(\bmv_{\bmtheta^\text{X},t}^\gamma\big)$. The importance weight $\bmsigma_\text{SM}^k\big(\bmv_{\bmtheta^\text{X},t}^\gamma\big)$ measures the relative effectiveness of the $k$th random realization $\bmh_{\leq t}^k$ of the hidden neurons in reproducing the desired behavior $\bmx_{\leq t}$ of the visible neurons.

\subsection{Communication Load}

As discussed, {\GEMSNN} requires bi-directional communication between all neurons and a CP. As seen in Fig.~\ref{fig:model-learning}, at each time $t$, unicast communication from neurons to CP is required in order to compute the importance weights by collecting information $\{\{\ell\big( x_{i,t}, \sigma(u_{i,t}^k)\big)\}_{k=1}^K\}_{i \in \set{X}}$ from all visible neurons. The resulting unicast communication load is $\CNtoCP = K|\set{X}|$ real numbers. 
The importance weights $\big\{\bmsigma_\text{SM}^k\big( \bmv_{\bmtheta^\text{X},t}^\gamma\big)\big\}_{k=1}^K$ are then sent back to all neurons, resulting a broadcast communication load from CP to neurons equal to $\CCPtoN = K (|\set{X}|+|\set{H}|)$ real numbers. 
As {\GEMSNN} requires computation of $K$ importance weights at CP, the communication loads increase linearly to the number $K$ of samples used for training.

\section{Experiments} \label{sec:experiments}

In this section, we evaluate the performance of the proposed inference and learning schemes on memorization and classification tasks defined on the neuromorphic data set MNIST-DVS \cite{serrano2015poker}. We conduct experiments by varying the number $K_I$ and $K$ of samples used for inference and training, and evaluate the performance in terms of robust decision making, uncertainty quantification, log-likelihood, classification accuracy, number of spikes \cite{merolla2014million}, communication load, and calibration \cite{guo2017calibration}.


The MNIST-DVS data set \cite{serrano2015poker} was generated by displaying slowly-moving handwritten digit images from the MNIST data set on an LCD monitor to a DVS neuromorphic camera \cite{lichtsteiner2006128}. For each pixel of an image, positive or negative events are recorded by the camera when the pixel's luminosity respectively increases or decreases by more than a given amount, and no event is recorded otherwise. In this experiment, images are cropped to $26 \times 26$ pixels, and uniform downsampling over time is carried out to obtain $T = 80$ time samples per each image as in \cite{zhao2014feedforward, henderson2015spike}. The training data set is composed of $900$ examples per each digit, from $0$ to $9$, and the test data set is composed of $100$ examples per digit. The signs of the spiking signals are discarded, producing a binary signal per pixel as in e.g., \cite{zhao2014feedforward}.

Throughout this section, as illustrated in Fig.~\ref{fig:model-inference}, we consider a generic, non-optimized, network architecture characterized by a set of $|\set{H}|$ fully connected hidden neurons, all receiving the exogeneous inputs as pre-synaptic signals, and a read-out visible layer, directly receiving pre-synaptic signals from all exogeneous inputs and all hidden neurons without recurrent connections between visible neurons. For synaptic and somatic filters, following the approach \cite{pillow2008spatio}, we choose a set of two or three raised cosine functions with a synaptic duration of $10$ time steps as synaptic filters, and, for somatic filter, we choose a single raised cosine function with duration of $10$ time steps.

Since we aim at demonstrating the advantages of using multiple samples for inference and learning, we only provide comparisons with the counterpart probabilistic methods that use a single sample in \cite{brea2013matching, jimenez2014stochastic, jang19:spm}. The advantages of probabilistic SNN models trained with conventional single-sample strategies over deterministic SNN models trained with the state-of-the-art method DECOLLE \cite{kaiser2020decolle} have been investigated in \cite{jang20:vowel, skatchkovsky2020reviewpt2}. It is shown therein that probabilistic methods can have significant gain in terms of accuracy in resource-constrained environments characterized by a small number of neurons and/or a short presentation time, even with a single sample. We do not repeat these experiments here.

\vspace{-0.15cm}
\subsection{Multi-Sample Learning on Memorization Task}

\begin{figure*}[t!]
\centering
\hspace{-0.3cm}
\subfigure[]{
\includegraphics[height=0.28\columnwidth]{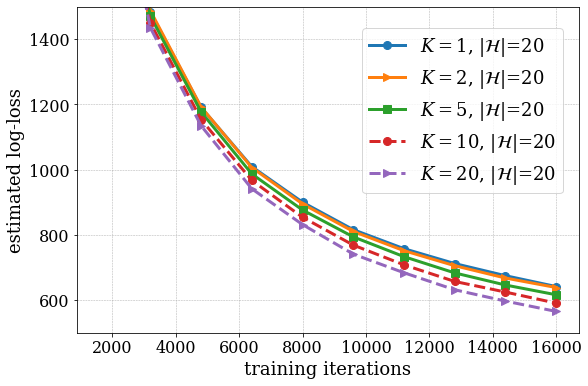} \label{fig:single-ll-train}
}
\hspace{-0.4cm}
\subfigure[]{
\includegraphics[height=0.28\columnwidth]{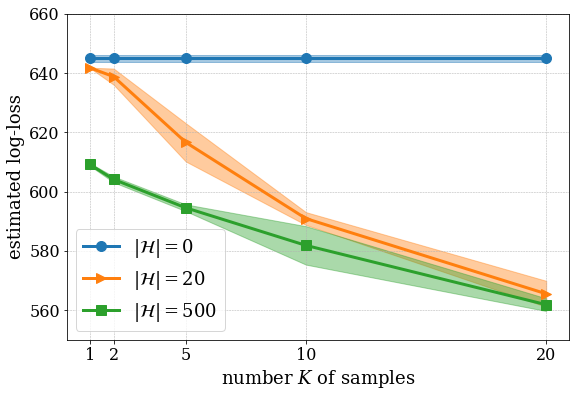} \label{fig:single-spike-K}
}
\hspace{-0.4cm}
\subfigure[]{
\includegraphics[height=0.28\columnwidth]{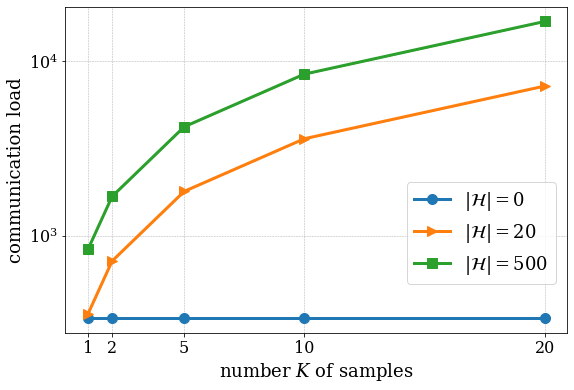} \label{fig:single-ll-K}
}
\hspace{-0.4cm}
\subfigure[]{
\includegraphics[height=0.28\columnwidth]{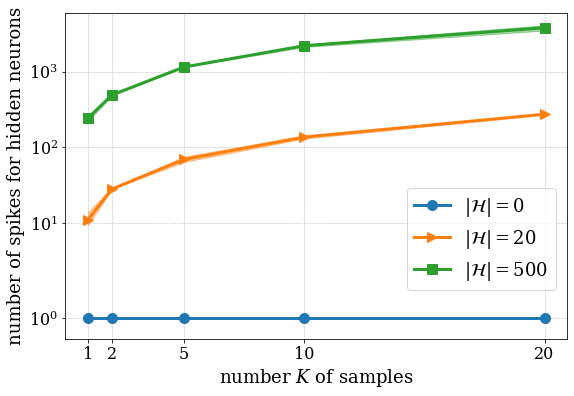} \label{fig:single-comm-K}
}
\vspace{-0.2cm}
\caption{Structured output memorization task trained on a single MNIST-DVS data point using {\GEMSNN}: (a) (Estimated) log-loss of the desired output as a function of the number of processed time samples for different values of the number $K = 1,2,5,10,20$ of samples with $|\set{H}| = 20$. (b) (Estimated) log-loss for the desired output; (c) broadcast communication load $\CCPtoN$ from CP to neurons; and (d) number of spikes emitted by the hidden neurons per unit time during training as a function of the number $K$ of samples in training for different values of $|\set{H}| = 20, 500$. Also shown for reference is the performance with $|\set{H}| = 0$ and shaded areas represent $95\%$ confidence intervals. }
\label{fig:single-training}
\vspace{-0.4cm}
\end{figure*}

We first focus on a structured output memorization task, which involves predicting the $13 \times 26$ spiking signals encoding the lower half of an MNIST-DVS digit from the $13 \times 26$ spiking signals encoding its top half. The top-half spiking signals are given as exogeneous inputs to the SNN, while the lower-half spiking signals are assigned as the desired outputs $\bmx_{\leq T}$ of the $|\set{X}| = 13 \times 26 = 338$ visible neurons.

For this memorization task, the accuracy of an SNN model with parameter $\bmtheta$ on a desired output signal $\bmx_{\leq T}$ is measured by the marginal log-loss $-\log p_{\bmtheta}(\bmx_{\leq T})$ obtained from \eqref{eq:ml-marginal}. The log-loss is estimated via the empirical average of the cross-entropy loss of the visible neurons over $20$ independent realizations of the hidden neurons \cite{simeone2018brief}. 

At a first example, we train an SNN model using {\GEMSNN} by presenting $200$ consecutive times a single MNIST-DVS training data point $\bmx_{\leq T}$, yielding a sequence of $200 \times T = 16000$ time samples. The trained SNN is tested on the same image, hence evaluating the capability of the SNN for memorization \cite{brea2013matching}. For training, hyperparameters such as learning rate $\eta$ and time constant $\gamma$ have been selected after a non-exhaustive manual search and are shared among all experiments. The initial learning rate $\eta = 5 \times 10^{-4}$ is decayed as $\eta = \eta/(1+0.2)$ every $40$ presentations of the data point, and we set $\gamma = 0.9$.

To start, Fig.~\ref{fig:single-ll-train} shows the evolution of the estimated log-loss of the desired output (the lower-half image) as more samples are processed by the proposed {\GEMSNN} rule for different values of the number $K = 1,2,5,10,20$ of samples and for $|\set{H}| = 0, 20, 500$ hidden neurons. The corresponding estimated log-loss at the end of training is illustrated as a function of $K$ in Fig.~\ref{fig:single-ll-K}. It can be observed that using more samples $K$ improves the training performance due to the optimization of an increasingly more accurate learning criterion. Improvements are also observed with an increasing number $|\set{H}|$ of hidden neurons. However, it should be emphasized that increasing the number of hidden neurons increases proportionally the number of weights in the SNN, while a larger $K$ does not increase the complexity of the model in terms of number of weights.

We now turn our attention to the number of spikes produced by the hidden neurons during training and to the requirements of the SNN trained with {\GEMSNN} in terms of the communication load. 
A larger $K$ implies a proportionally larger number of spikes emitted by the hidden neurons during training, as seen in Fig.~\ref{fig:single-spike-K}. We recall that the number of spikes is a proxy for the energy consumption of the SNN. Furthermore, Fig.~\ref{fig:single-comm-K} shows the broadcast communication load $\CCPtoN$ from CP to neurons as a function of $K$ for different values of $|\set{H}|$. The communication load of an SNN trained using {\GEMSNN} increases linearly with $K$ and $|\set{H}|$. 
From Fig.~\ref{fig:single-training}, the proposed {\GEMSNN} is seen to enable a flexible trade-off between communication load and energy consumption, on the one hand, and training performance, on the other, by leveraging the availability of $K$ samples.

\vspace{-0.15cm}

\subsection{Multi-Sample Inference on Classification Task}

Next, we consider a handwritten digit classification task based on the MNIST-DVS data set. In order to evaluate the advantage of multi-sample inference rule proposed in Sec.~\ref{sec:multi-inference}, we focus on a binary classification task. To this end, we consider a probabilistic SNN with $|\set{X}| = 2$ visible output neurons in the read-out layer, one for each class of two digits `$0$' and `$1$'. The $26 \times 26$ spiking signals encoding an MNIST-DVS image are given as exogeneous inputs to the SNN, while the digit labels $\{0,1\}$ are encoded by the neurons in the read-out layer. The output neuron $c \in \set{X}$ corresponding to the correct label is assigned a desired output spike signals $x_{c,t} = 1$ for $t=1,\ldots,T$, while the other neuron $c' \neq c$ are assigned $x_{c',t} = 0$ for $t=1,\ldots,T$.

We train an SNN model with $|\set{H}|=4$ hidden neurons on the $1800$ training data points for digits $\{0,1\}$ by using the proposed learning rule {\GEMSNN} with $K=5$ samples. We set the constant learning rate $\eta = 10^{-4}$ and time constant $\kappa = 0.2$, which have been selected after a non-exhaustive manual search. For testing on the $200$ test data points, we implement a majority rule with $K_I$ spiking signals $\{\bmx_{\leq T}^k\}_{k=1}^{K_I}$ for inference. The $K_I$ samples are used to obtain the final classification decisions $\hat{c}$ using \eqref{eq:rate-class}-\eqref{eq:majority}; and to quantify the aleatoric uncertainty of the decisions using \eqref{eq:majority-prob}. In particular, the probability of each class is estimated with the empirical average as in \eqref{eq:majority-prob}.

\begin{figure*}[t!]
\centering
\hspace{-0.3cm}
\subfigure[]{
\includegraphics[height=0.31\columnwidth]{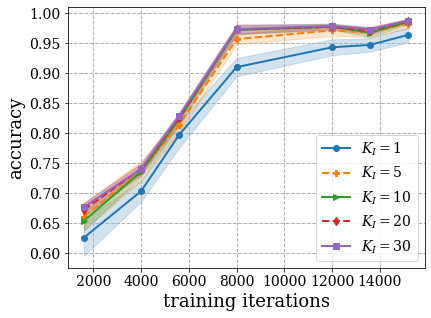} \label{fig:inference-acc}
}
\hspace{-0.4cm}
\subfigure[]{
\includegraphics[height=0.31\columnwidth]{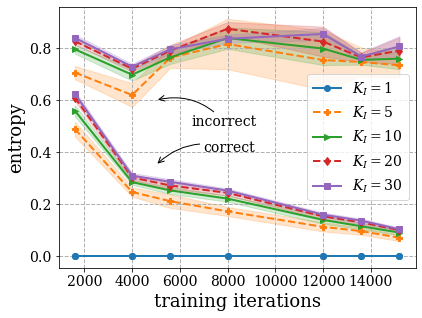}
\label{fig:inference-entropy}
}
\vspace{-0.2cm}
\caption{Classification task trained on MNIST-DVS data points of two digits $\{0,1\}$ using {\GEMSNN} with $K=5$ samples: (a) Classification accuracy; and (b) Average entropy of classes measured for correct and incorrect decisions as a function of processed time samples for different values $K_I=1,5,10,20,30$ of samples in inference. The shaded areas represent $95\%$ confidence intervals.}
\label{fig:multi-inference}
\vspace{-0.4cm}
\end{figure*}

To start, we plot in Fig.~\ref{fig:inference-acc} the classification accuracy on test data set as a function of the number of training iterations. A larger $K_I$ is seen to improve the test accuracy thanks to the more robust final decisions obtained by the majority rule \eqref{eq:majority}. As discussed in \eqref{eq:majority-error}, the probability of error of the majority rule decreases exponentially with $K_I$. In particular, after training on the $100$ data points, yielding a sequence of $8000$ time samples, the test accuracy obtained by single-sample ($K_I = 1$) inference is $90.9\%$, which is improved to $97.2\%$ with $K_I = 20$ samples. 

Next, we evaluate the uncertainty of the predictions as a function of the number $K_I$ of samples used for inference. Using $K_I$ decisions, the probability assigned to each class by the trained model is given by \eqref{eq:majority-prob}. Accordingly, each class is assigned a degree of confidence that depends on the fraction of the $K_I$ decisions assigned to it. 
In order to evaluate the uncertainty in this decision, we evaluate the entropy of the distribution \eqref{eq:majority-prob} across the classes. More precisely, we separately average the entropy for correct and incorrect decisions. In this way, we can assess the capacity of the model to quantify uncertainty for both correct and incorrect decisions. 
As seen in Fig.~\ref{fig:inference-entropy}, when $K_I=1$, the entropy is zero for both classes, since the model is unable to quantify uncertainty. In contrast, with $K_I > 1$, the entropy of correct decisions decreases as a function of training iterations, while the entropy of incorrect decisions remains large, i.e., close to $1$ bit. This indicates that the model reports well calibrated decisions.

\subsection{Multi-Sample Learning on Classification Task}

\begin{figure}[t!]
\centering
\includegraphics[height=0.36\columnwidth]{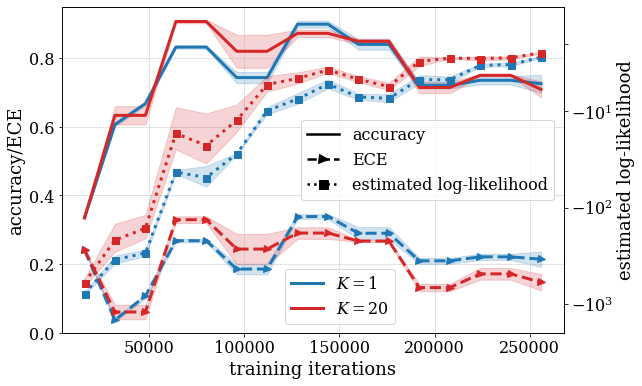} 
\vspace{-0.2cm}
\caption{Classification task trained on MNIST-DVS data points of digits $\{0,1,2\}$: 
Estimated log-likelihood, classification accuracy and expected calibration error (ECE) \eqref{eq:ece} of test data points as a function of processed time samples for different values $K$ of samples in training using {\GEMSNN}. The accuracy and ECE are measured using $K_I = 2$ samples. The shaded areas represent $95\%$ confidence intervals.
}
\label{fig:general-ll-ece-gem} 
\vspace{-0.4cm}
\end{figure}

Finally, we investigate the advantage of using multiple samples for training in a classification task. We consider a SNN with $|\set{X}| =3$ visible output neurons in the read-out layer, one for each class of three digits $\{0,1,2\}$. We train the SNN with $|\set{H}| = 200$ hidden neurons on the $2700$ training data points of digits $\{0,1,2\}$ by using {\GEMSNN}, and the SNN is tested on the $300$ test data points. 
For testing, we adopt the majority rule \eqref{eq:rate-class}-\eqref{eq:majority} to evaluate accuracy. We also consider {\em calibration} as a performance metric following \cite{guo2017calibration}. To this end, the prediction probability $\hat{p}$, or confidence, of a decision is derived from the vote count $\bmz = (z_c : c \in \set{X})$, with $z_c = \sum_{k=1}^{K_I} \mathds{1}(\bmx_{\leq T}^k \in \mathds{X}_c)$, using the SoftMax function, i.e., 
\begin{align*}
    \hat{p} = \sigma_{\text{SM}}^{\hat{c}}\big( \bmz \big).    
\end{align*}
The expected calibration error (ECE) measures the difference in expectation between confidence and accuracy, i.e., \cite{guo2017calibration}
\begin{align} \label{eq:ece}
    \text{ECE} = \mathbb{E}_{\hat{p}} \Big[ \big| \mathbb{P}\big( \hat{c} = c | \hat{p} = p \big) - p \big| \Big].
\end{align}
In \eqref{eq:ece}, the probability $\mathbb{P}\big( \hat{c} = c| \hat{p}=p\big)$ is the probability that $\hat{c}$ is the correct decision for inputs that yield accuracy $\hat{p} = p$. The ECE can be estimated by using quantization and empirical averages as detailed in \cite{guo2017calibration}.

We plot in Fig.~\ref{fig:general-ll-ece-gem} the test accuracy, ECE, and estimated log-likelihood of the desired output spikes as a function of the number of training iterations. A larger $K$ is seen to improve the test log-likelihood thanks to the optimization of a more accurate training criterion. The larger log-likelihood also translates into a model that more accurately reproduces conditional probability of outputs given inputs \cite{guo2017calibration, bishop1994mixture}, which in turn enhances calibration. In contrast, accuracy can be improved with a larger $K$ but only if regularization via early stopping is carried out. This points to the fact that the goal of minimizing the log-loss of specific desired output spiking signals is not equivalent to minimizing the classification error \cite{guo2017calibration}. More results can be found in Appendix, when we compare the performance with several alternative training schemes based on multiple samples.

\section{Conclusions} \label{sec:conclusion}

In this paper, we have explored the implications for inference and learning of one of the key unique properties of probabilistic spiking neural models as opposed to standard deterministic models, namely their capacity to generate multiple independent spiking outputs for any input. 
We have introduced inference and online learning rules that can leverage multiple samples through majority rule and generalized expectation-maximization (GEM), respectively. 
The multi-sample inference rule has the advantage of robustifying decisions and of quantifying uncertainty. 
The GEM-based learning rule is derived from an estimation of the log-loss via importance sampling. 
Experiments on the neuromorphic data set MNIST-DVS have demonstrated that multi-sample inference and learning rules, as compared to conventional single-sample schemes, can improve training and test performance in terms of accuracy and calibration.

While this work considered the log-loss of specific desired output spiking signals as the learning criterion, similar rules can be derived by considering other reward functions, such as Van Rossum (VR) distance \cite{zenke2018superspike, rossum2001novel}. The multi-sample algorithms for probabilistic SNN models proposed in this paper can be also extended to networks of spiking Winner-Take-All (WTA) circuits \cite{jang20:vowel, mostafa2018learning}, which process multi-valued spikes.

\appendices
\section{Alternative Multi-Sample Strategies}

In this section, we review two alternative learning rules that tackle the ML problem \eqref{eq:ml} by leveraging $K$ independent samples from the hidden neurons. We also present a performance comparison on the MNIST-DVS classification task studied in Sec.~\ref{sec:experiments}.

\subsection{{\MBSNN}: Mini-Batch Online Learning for SNNs}

The first scheme is a direct mini-batch extension of the online learning rule for SNN developed in \cite{brea2013matching, jimenez2014stochastic, jang19:spm}, which we term {\MBSNN}. The learning algorithm for SNN introduced in \cite{brea2013matching, jimenez2014stochastic, jang19:spm} aims at minimizing the upper bound of the log-loss in \eqref{eq:general-elbo} obtained via Jensen's inequality as 
\begin{align} \label{eq:elbo-snn}
    - \log p_{\bmtheta}(\bmx_{\leq T}) &\leq -\mathbb{E}_{ p_{\bmtheta^\text{H}}(\bmh_{\leq T}||\bmx_{\leq T})} \Big[ \log \frac{p_{\bmtheta}(\bmx_{\leq T}, \bmh_{\leq T})}{p_{\bmtheta^\text{H}}(\bmh_{\leq T}||\bmx_{\leq T})} \Big] \cr 
    &= \mathbb{E}_{p_{\bmtheta^\text{H}}(\bmh_{\leq T}||\bmx_{\leq T})}\Big[ \underbrace{ \sum_{t=1}^T \sum_{i \in \set{X}} \ell\big( x_{i,t}, \sigma(u_{i,t})\big)}_{=~ -v_{\bmtheta^\text{X},T}:~ \text{learning signal} } \Big] := \set{L}_{\bmx_{\leq T}}(\bmtheta),
\end{align}
where we have recalled the notation $v_{\bmtheta^\text{X},T} = \log p_{\bmtheta^\text{X}}(\bmx_{\leq T}||\bmh_{\leq T-1})$ in \eqref{eq:gem-log-weight-ls}. The minimization of the bound $\set{L}_{\bmx_{\leq T}}(\bmtheta)$ via SGD in the direction of a stochastic estimate of the gradient $-\grad_{\bmtheta} \set{L}_{\bmx_{\leq T}}(\bmtheta)$ results the update rule $\bmtheta \leftarrow \bmtheta - \eta \cdot \grad_{\bmtheta} \set{L}_{\bmx_{\leq T}}(\bmtheta)$, with $\eta > 0$ being the learning rate. The gradient is estimated in \cite{jang19:spm} by using the REINFORCE gradient \cite{simeone2018brief}
\begin{align}
    &\grad_{\bmtheta} \set{L}_{\bmx_{\leq T}}(\bmtheta) = \mathbb{E}_{p_{\bmtheta^\text{H}}(\bmh_{\leq T}|| \bmx_{\leq T})}\Big[ -\grad_{\bmtheta}v_{\bmtheta^\text{X},T} - v_{\bmtheta^\text{X},T} \cdot \grad_{\bmtheta} \log p_{\bmtheta^\text{H}}(\bmh_{\leq T}||\bmx_{\leq T}) \Big] \cr 
    &\quad = \mathbb{E}_{p_{\bmtheta^\text{H}}(\bmh_{\leq T}|| \bmx_{\leq T})}\Big[ \sum_{t=1}^T \sum_{i \in \set{X}} \grad_{\bmtheta}\ell\big( x_{i,t},\sigma(u_{i,t})\big) + v_{\bmtheta^\text{X},T} \cdot \sum_{t=1}^T \sum_{i \in \set{H}} \grad_{\bmtheta} \ell\big(h_{i,t},\sigma(u_{i,t})\big) \Big].
\end{align}

An MC estimate of the gradient can be obtained by drawing a mini-batch of $K$ independent spiking signals $\bmh_{\leq T}^{1:K}=\{\bmh_{\leq T}^k\}_{k=1}^K$ of the hidden neurons via $K$ independent forward passes through the SNN, i.e., $\bmh_{\leq T}^k \sim p_{\bmtheta^\text{H}}(\bmh_{\leq T}^k||\bmx_{\leq T})$ for $k=1,\ldots,K$, and by evaluating for visible neuron $i \in \set{X}$,
\begin{subequations} \label{eq:mc-grad-batch}
\begin{align} \label{eq:mc-grad-batch-vis}
    \grad_{\bmtheta_i} \set{L}^K_{\bmx_{\leq T}}(\bmtheta) := \frac{1}{K} \sum_{k=1}^K \sum_{t=1}^T \grad_{\bmtheta_i} \ell\big( x_{i,t},\sigma(u_{i,t}^k)\big),
\end{align} 
and for hidden neuron $i \in \set{H}$,
\begin{align} \label{eq:mc-grad-batch-hid}
    &\grad_{\bmtheta_i} \set{L}^K_{\bmx_{\leq T}}(\bmtheta) :=  \frac{1}{K} \sum_{k=1}^K \bigg(  \big( v_{\bmtheta^\text{X},T}^k - \bmb_{i,T}^k\big) \cdot  \sum_{t=1}^T \grad_{\bmtheta_i} \ell\big( h_{i,t}^k,\sigma(u_{i,t}^k)\big) \bigg), 
\end{align}
\end{subequations}
In \eqref{eq:mc-grad-batch}, we have introduced baseline, also known as control variates, signals $\bmb_{i,T}^k$ for each $k$th sample of mini-batch as means to reduce the variance of the gradient estimator. Following the approach in \cite{peters2008reinforcement}, an optimized baseline can be evaluated as 
\begin{align} \label{eq:mb-baseline}
    \bmb_{i,T}^k = \frac{\mathbb{E}\Big[ v_{\bmtheta^\text{X},T}^k \cdot \Big(\sum_{t=1}^T \grad_{\bmtheta_i} \ell\big( h_{i,t}^k,\sigma(u_{i,t}^k)\big)\Big)^2 \Big]}{ \mathbb{E} \Big[ \Big(\sum_{t=1}^T \grad_{\bmtheta_i} \ell\big( h_{i,t}^k,\sigma(u_{i,t}^k)\big)\Big)^2 \Big]}.
\end{align}

In order to obtain an online rule, the model parameters $\bmtheta$ are updated at each time $t$ based on the data $\bmx_{\leq t}$ via SGD by using the discounted version of the gradient estimator $\grad_{\bmtheta} \set{L}^K_{\bmx_{\leq T}}(\bmtheta)$ in \eqref{eq:mc-grad-batch}, which is given as for visible neuron $i \in \set{X}$,
\begin{align*} 
    \grad_{\bmtheta_i} \set{L}_{\bmx_{\leq t}}^{K,\gamma}(\bmtheta) := \frac{1}{K} \sum_{k=1}^K \sum_{t'=0}^{t-1} \gamma^{t'} \grad_{\bmtheta_i} \ell\big( x_{i,t-t'}, \sigma(u_{i,t-t'}^k)\big),
\end{align*}
and for hidden neuron $i \in \set{H}$,
\begin{align*} 
    &\grad_{\bmtheta_i} \set{L}_{\bmx_{\leq t}}^{K,\gamma}(\bmtheta) := \frac{1}{K} \sum_{k=1}^K \bigg( \big( v_{\bmtheta^\text{X},t}^{k,\gamma} - \bmb_{i,t}^{k,\gamma}\big) \cdot \sum_{t'=0}^{t-1} \gamma^{t'} \grad_{\bmtheta_i} \ell\big( h_{i,t-t'}^k, \sigma(u_{i,t-t'}^k)\big) \bigg),
\end{align*}
where $\gamma \in (0,1)$ is a discount factor.

In a similar manner to {\GEMSNN}, the batch processing required in learning signal $v_{\bmtheta^\text{X},T}^k$ and the baseline $\bmb_{i,T}^k$ can be addressed by computing the discounted values $v_{\bmtheta^\text{X},t}^{k,\gamma}$ from \eqref{eq:gem-log-weight} and the corresponding baseline $\bmb_{i,t}^{k,\gamma}$ using temporal average operator. The resulting online learning rule, which we refer to Mini-batch online learning for SNNs ({\MBSNN}), updates the parameters $\bmtheta_i$ of each neuron $i$ in the direction of the MC estimate of the gradient $-\grad_{\bmtheta_i} \set{L}_{\bmx_{\leq t}}^{K,\gamma}(\bmtheta)$, yielding the update rule at time $t$ as follows: for visible neuron $i \in \set{X}$,
\begin{subequations} \label{eq:mb-update}
\begin{align} \label{eq:mb-update-vis}
    \Delta \bmtheta_i = \frac{1}{K} \sum_{k=1}^K  \Big\langle \grad_{\bmtheta_i} \ell\big( x_{i,t},\sigma(u_{i,t}^k)\big) \Big\rangle_{\gamma}, 
\end{align}
and for hidden neuron $i \in \set{H}$,
\begin{align} \label{eq:mb-update-hid}
    \Delta \bmtheta_i = \frac{1}{K} \sum_{k=1}^K \Big( v_{\bmtheta^\text{X},t}^{k,\gamma} - \bmb_{i,t}^{k,\gamma} \Big) \cdot \Big\langle \grad_{\bmtheta_i} \ell\big( h_{i,t}^k,\sigma(u_{i,t}^k)\big) \Big\rangle_{\gamma},
\end{align}
\end{subequations}

We note that learning rules based on the conventional single-sample estimator in \cite{jimenez2014stochastic, brea2013matching, jang19:spm} rely on a single stochastic sample $\bmh_{\leq t}$ for the hidden neurons. It has a generally high variance, only partially decreased by the presence of the baseline control variates. By leveraging a mini-batch of $K$ samples, it is anticipated that {\MBSNN} with $K>1$ can potentially improve the learning performance by reducing the variance of the single-sample estimator.


\subsection{{\IWSNN}: Importance-Weighted Multi-Sample Online Learning for SNNs}

The second alternative approach uses the available $K$ independent samples to obtain an increasingly more accurate bound on the log-loss $\log p_{\bmtheta}(\bmx_{\leq T})$ in \eqref{eq:ml}. The approach adapts the principles of the importance weighting method, which are introduced in \cite{mnih2016variational, burda2015importance, domke2018iwvi} for conventional probabilistic models, to probabilistic SNNs.

We start by considering an unbiased, importance-weighted estimator of the marginal $p_{\bmtheta}(\bmx_{\leq T})$ of the form
\begin{align} \label{eq:iw-estimator}
    p_{\bmtheta}(\bmx_{\leq T}) \approx \frac{1}{K} \sum_{K=1}^K \frac{p_{\bmtheta}(\bmx_{\leq T},\bmh_{\leq T}^k)}{ p_{\bmtheta^\text{H}}(\bmh_{\leq T}^k||\bmx_{\leq T})} &= \frac{1}{K} \sum_{k=1}^K p_{\bmtheta^\text{X}}(\bmx_{\leq T}||\bmh_{\leq T-1}^k) \cr 
    &= \frac{1}{K}\sum_{k=1}^K \exp\big( v_{\bmtheta^\text{X},T}^k\big) := R_{\bmtheta^\text{X},T}^K,
\end{align}
where $\bmh_{\leq T}^{1:K}$ are $K$ independent spiking signals of the hidden neurons drawn from the causally conditioned distribution $p_{\bmtheta^\text{H}}(\bmh_{\leq T}^{1:K}||\bmx_{\leq T}) = \prod_{k=1}^K p_{\bmtheta^\text{H}}(\bmh_{\leq T}^k||\bmx_{\leq T})$ in \eqref{eq:causally-cond}, and we have used the notation $v_{\bmtheta^\text{X},T}^k = \log p_{\bmtheta^\text{X}}(\bmx_{\leq T}||\bmh_{\leq T-1}^k)$ in \eqref{eq:gem-log-weight-ls}. The estimator $R_{\bmtheta^\text{X},T}^K$ in \eqref{eq:iw-estimator} for the marginal is unbiased, i.e., $\mathbb{E}_{p_{\bmtheta^\text{H}}(\bmh_{\leq T}^{1:K}||\bmx_{\leq T})}[R_{\bmtheta^\text{X},T}^K] = p_{\bmtheta}(\bmx_{\leq T})$, and can be used to obtain the upper bound of the log-loss as
\begin{align} \label{eq:iw-elbo-T}
    -\log p_{\bmtheta}(\bmx_{\leq T}) &= -\log \mathbb{E}_{p_{\bmtheta^\text{H}}(\bmh_{\leq T}^{1:K}||\bmx_{\leq T})} \big[ R_{\bmtheta^\text{X},T}^K \big] \cr
    &\leq - \mathbb{E}_{p_{\bmtheta^\text{H}}(\bmh_{\leq T}^{1:K}||\bmx_{\leq T})}\Big[ \log R_{\bmtheta^\text{X},T}^K \Big] := \set{L}_{\bmx_{\leq T}}^K(\bmtheta). 
\end{align}
We note that for $K \rightarrow \infty$, the upper bound $\set{L}_{\bmx_{\leq T}}^K(\bmtheta)$ in \eqref{eq:iw-elbo-T} converges to the exact log-loss $-\log p_{\bmtheta}(\bmx_{\leq T})$ \cite{burda2015importance}.

The minimization of the bound $\set{L}_{\bmx_{\leq T}}^K(\bmtheta)$ in \eqref{eq:iw-elbo-T} via SGD in the direction of the gradient $-\grad_{\bmtheta} \set{L}_{\bmx_{\leq T}}^K(\bmtheta)$ yields the update rule $\bmtheta \leftarrow \bmtheta - \eta \cdot \grad_{\bmtheta} \set{L}_{\bmx_{\leq T}}^K(\bmtheta)$ with $\eta>0$ the learning rate. The gradient is obtained from the REINFORCE gradient method as
\begin{align} \label{eq:iw-grad-T}
    &\grad_{\bmtheta} \set{L}_{\bmx_{\leq T}}^K(\bmtheta) = - \mathbb{E}_{p_{\bmtheta^\textH}(\bmh_{\leq T}^{1:K}||\bmx_{\leq T})} \Big[ \grad_{\bmtheta} \log R_{\bmtheta^\text{X},T}^K + \log R_{\bmtheta^\text{X},T}^K \cdot \grad_{\bmtheta} \log p_{\bmtheta^\text{H}}(\bmh_{\leq T}^{1:K}||\bmx_{\leq T}) \Big],
\end{align}
where the first gradient in \eqref{eq:iw-grad-T} can be computed as
\begin{align}
    \grad_{\bmtheta} \log R_{\bmtheta^\text{X},T}^K &= \frac{1}{R_{\bmtheta^\text{X},T}^K} \cdot \grad_{\bmtheta}R_{\bmtheta^\text{X},T}^K \cr 
    &= \frac{1}{ \sum_{k'=1}^K \exp\big( v_{\bmtheta^\text{X},T}^{k'}\big) } \cdot \sum_{k=1}^K \exp\big( v_{\bmtheta^\text{X},T}^k\big) \cdot \grad_{\bmtheta} v_{\bmtheta^\text{X},T}^k \cr 
    &= -\sum_{k=1}^K \bmsigma_{\text{SM}}^k\Big( \bmv_{\bmtheta^\text{X},T}\Big) \cdot \sum_{t=1}^T \sum_{i \in \set{X}} \grad_{\bmtheta} \ell\big( x_{i,t},\sigma(u_{i,t}^k)\big)
\end{align}
by using the chain rule, while the second gradient in \eqref{eq:iw-grad-T} can be computed as
\begin{align}
    \grad_{\bmtheta} \log p_{\bmtheta^\text{H}}(\bmh_{\leq T}^{1:K}||\bmx_{\leq T}) = - \sum_{k=1}^K \sum_{t=1}^T \sum_{i \in \set{H}} \grad_{\bmtheta} \ell\big( h_{i,t}^k,\sigma(u_{i,t}^k)\big).
\end{align}
As a result, we have the MC estimate of the gradient using the $K$ samples $\bmh_{\leq T}^{1:K}$ as: for visible neuron $i \in \set{X}$,
\begin{align} \label{eq:iw-grad-T-vis}
    \grad_{\bmtheta_i} \set{L}_{\bmx_{\leq T}}^K(\bmtheta) = \sum_{k=1}^K \bigg( \bmsigma_{\text{SM}}^k\Big( \bmv_{\bmtheta^\text{X},T}\Big) \cdot \sum_{t=1}^T \grad_{\bmtheta_i} \ell\big(x_{i,t},\sigma(u_{i,t}^k)\big) \bigg),
\end{align}
and for hidden neuron $i \in \set{H}$,
\begin{align} \label{eq:iw-grad-T-hid}
    &\grad_{\bmtheta_i} \set{L}_{\bmx_{\leq T}}^K(\bmtheta) = \Big( \log R_{\bmtheta^\text{X},T}^K - \bmb_{i,T} \Big) \cdot \sum_{k=1}^K \sum_{t=1}^T \grad_{\bmtheta_i} \ell\big(h_{i,t}^k,\sigma(u_{i,t}^k)\big),
\end{align}
where we have introduced baseline signals $\bmb_{i,T}$ to minimize the variance of the gradient estimator following similar way in \eqref{eq:mb-baseline}.

In order to obtain an online rule, the model parameters $\bmtheta$ are updated at each time $t$ via SGD by using the discounted version of the gradient estimator $\grad_{\bmtheta} \set{L}_{\bmx_{\leq T}}^K(\bmtheta)$ in \eqref{eq:iw-grad-T}, given as: for visible neuron $i \in \set{X}$,
\begin{subequations} \label{eq:iw-grad}
\begin{align} \label{eq:iw-grad-vis}
    \grad_{\bmtheta_i} \set{L}_{\bmx_{\leq t}}^{K,\gamma}(\bmtheta) &=  \sum_{k=1}^K \bigg( \bmsigma_{\text{SM}}^k\Big( \bmv_{\bmtheta^\text{X},t}^{\gamma}\Big) \cdot \sum_{t'=0}^{t-1} \gamma^{t'} \grad_{\bmtheta_i} \ell\big( x_{i,t-t'},\sigma(u_{i,t-t'}^k)\big) \bigg),
\end{align}
and for hidden neuron $i \in \set{H}$, 
\begin{align} \label{eq:iw-grad-hid}
    \grad_{\bmtheta_i} \set{L}_{\bmx_{\leq t}}^{K,\gamma}(\bmtheta) &= \Big( \log R_{\bmtheta^\text{X},t}^{K,\gamma} - \bmb_{i,t}^{\gamma}\Big) \cdot \sum_{k=1}^K \sum_{t'=0}^{t-1} \gamma^{t'} \grad_{\bmtheta_i} \ell\big( h_{i,t-t'}^k,\sigma(u_{i,t-t'}^k)\big), 
\end{align}
\end{subequations}
where $\gamma \in (0,1)$ is a discount factor. In a similar manner to {\GEMSNN} and {\MBSNN}, the batch processing required in importance weight $\bmsigma_{\text{SM}}^k\big(\bmv_{\bmtheta^\text{X},T}\big)$ and in importance-weighted estimator $R_{\bmtheta^\text{X},T}^K$, we use the discounted values $v_{\bmtheta^\text{X},t}^{k,\gamma}$ in \eqref{eq:gem-log-weight} to compute the discounted version of the estimator $R_{\bmtheta^\text{X},t}^{K,\gamma} = \frac{1}{K} \sum_{k=1}^K \exp(v_{\bmtheta^\text{X},t}^{k,\gamma})$ and the corresponding baseline $\bmb_{i,t}^{\gamma}$. 
The resulting SGD-based learning rule, which we refer to Importance-Weighted online learning for SNNs ({\IWSNN}), updates the parameters $\bmtheta_i$ of each neuron $i$ in the direction of the MC estimate of the gradient $-\grad_{\bmtheta_i} \set{L}_{\bmx_{\leq t}}^{K,\gamma}(\bmtheta)$ at time $t$ as follows: for visible neuron $i \in \set{X}$, 
\begin{subequations} \label{eq:iw-update}
\begin{align} \label{eq:iw-update-vis}
    \Delta \bmtheta_i = \sum_{k=1}^K \bmsigma_{\text{SM}}^k\Big( \bmv_{\bmtheta^\text{X},t}^{\gamma} \Big) \cdot \Big\langle \grad_{\bmtheta_i} \ell\big( x_{i,t},\sigma(u_{i,t}^k)\big) \Big\rangle_{\gamma},
\end{align}
and for hidden neuron $i \in \set{H}$, 
\begin{align} \label{eq:iw-update-hid}
    \Delta \bmtheta_i = \Big( \log R_{\bmtheta^\text{X},t}^{K,\gamma} - \bmb_{i,t}^{\gamma}\Big) \cdot \sum_{k=1}^K \Big\langle  \grad_{\bmtheta_i} \ell\big( h_{i,t}^k, \sigma(u_{i,t}^k)\big) \Big\rangle_{\gamma}.
\end{align}
\end{subequations}


\subsection{Interpreting {\MBSNN} and {\IWSNN}}

Following the discussion around \eqref{eq:three-factor}, the update rules {\MBSNN} and {\IWSNN} follow a standard three-factor format and is local, with the exception of the learning signals used for the update of hidden neurons' parameters. 

In {\MBSNN}, for each neuron $i \in \set{V}$, the gradient $\grad_{\bmtheta_i} \set{L}_{\bmx_{\leq t}}^{K,\gamma}(\cdot)$ in \eqref{eq:mb-update} contains the derivative of the cross-entropy loss $\grad_{\bmtheta_i} \ell\big( s_{i,t}^k,\sigma(u_{i,t}^k)\big)$, with $s_{i,t}^k = x_{i,t}$ for visible neuron $i \in \set{X}$ and $s_{i,t}^k = h_{i,t}^k$ for hidden neuron $i \in \set{H}$. As detailed in \eqref{eq:gem-update}, the derivative includes the synaptic filtered trace $\overra{s}_{j,t}^k$ of pre-synaptic neuron $j$; the somatic filtered trace $\overla{s}_{i,t}^k$ of post-synaptic neuron $i$; and the post-synaptic error $s_{i,t}^k - \sigma(u_{i,t}^k)$. Furthermore, for hidden neurons $\set{H}$, the per-sample global learning signals $\{v_{\bmtheta^\text{X},t}^{k,\gamma}\}_{k=1}^K$ in \eqref{eq:gem-log-weight} are commonly used to guide the update in \eqref{eq:mb-update-hid}, while the visible neurons $\set{X}$ do not use the learning signal. The common per-sample learning signals $\{v_{\bmtheta^\text{X},t}^{k,\gamma}\}_{k=1}^K$ can be interpreted as indicating how effective the current, randomly sampled, behavior $\bmh_{\leq t}^k$ of hidden neurons is ensuring the minimization of the log-loss of the desired observation $\bmx_{\leq t}$ for the visible neurons.

\begin{table}[t]
\caption{Learning Communication Loads (in real numbers)}
\label{tab:comparison}
\vspace{-0.3cm}
\begin{center}
\begin{small}
\begin{sc}
\begin{tabular}{lcccr}
\toprule
Scheme & Unicast $\CNtoCP$ & Broadcast $\CCPtoN$\\
\midrule
{\GEMSNN} & $K|\set{X}|$ & $K(|\set{X}| + |\set{H}|)$ \\
{\MBSNN} & $K|\set{X}|$ & $K|\set{H}|$ \\
{\IWSNN} & $K|\set{X}|$ & $K|\set{X}| + |\set{H}|$ \\
\bottomrule
\end{tabular}
\end{sc}
\end{small}
\end{center}
\vspace{-0.3cm}
\end{table}

In {\IWSNN}, for each neuron $i \in \set{V}$, the gradient $\grad_{\bmtheta_i} \set{L}_{\bmx_{\leq t}}^{K,\gamma}(\cdot)$ in \eqref{eq:iw-update} also contains the derivative of the cross-entropy loss $\grad_{\bmtheta_i} \ell\big( s_{i,t}^k,\sigma(u_{i,t}^k)\big)$. The dependence on the pre-synaptic filtered trace, post-synaptic somatic filtered trace, and post-synaptic error is applied for each sample $k$ as for {\GEMSNN} and {\MBSNN}, while the learning signal is given in different form to each neuron. As in \eqref{eq:iw-update-vis}, for visible neurons $\set{X}$, the importance weights $\{\bmsigma_{\text{SM}}^k\big( \bmv_{\bmtheta^\text{X},t}^{\gamma}\big)\}_{k=1}^K$ computed using the SoftMax function are given as the common learning signals, with the contribution of each sample being weighted by $\bmsigma_{\text{SM}}^k\big( \bmv_{\bmtheta^\text{X},t}^{\gamma}\big)$. As for {\GEMSNN}, the importance weight measures the relative effectiveness of the $k$th random realization $\bmh_{\leq t}^k$ of the hidden neurons in reproducing the desired behavior $\bmx_{\leq t}$ of the visible neurons. In contrast, for hidden neurons $\set{H}$, a scalar $\log R_{\bmtheta^\text{X},t}^{K,\gamma} = \frac{1}{K} \sum_{k=1}^K \exp\big( v_{\bmtheta^\text{X},t}^{k,\gamma}\big)$ is commonly used in \eqref{eq:iw-update-hid} as a global learning signal, indicating how effective their overall behavior across $K$ samples is in ensuring the minimization of the log-loss of observation $\bmx_{\leq t}$.

\begin{figure*}[ht!]
\centering
\hspace{-0.4cm}
\subfigure[]{
\includegraphics[height=0.28\columnwidth]{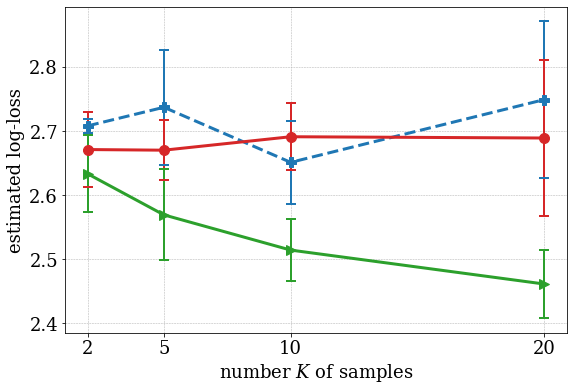} \label{fig:general-comp-ll}
}
\hspace{-0.4cm}
\subfigure[]{
\includegraphics[height=0.28\columnwidth]{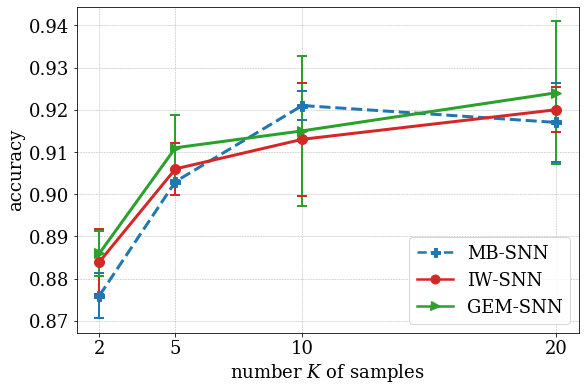} \label{fig:general-comp-acc}
}
\hspace{-0.4cm}
\subfigure[]{
\includegraphics[height=0.28\columnwidth]{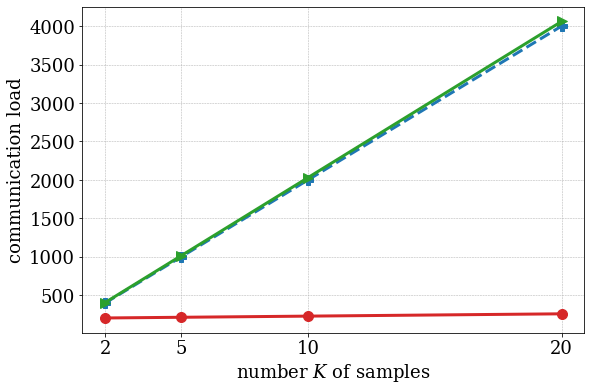} \label{fig:general-comp-comm}
}
\hspace{-0.4cm}
\subfigure[]{
\includegraphics[height=0.28\columnwidth]{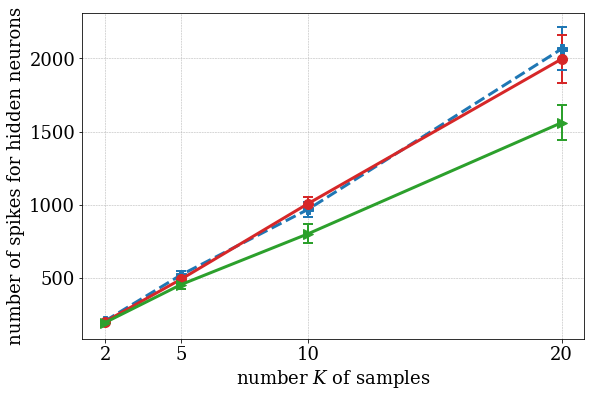} \label{fig:general-comp-spike}
}
\vspace{-0.1cm}
\caption{Classification task trained on MNIST-DVS data set using {\MBSNN}, {\IWSNN} and {\GEMSNN} versus $K$, with $95\%$ confidence intervals accounting for the error bars: (a) (estimated) log-loss for the desired output on test data set; (b) classification accuracy of the test data set; (c) broadcast communication load $\CCPtoN$ from CP to neurons in Table~\ref{tab:comparison}; and (d) number of spikes emitted by the hidden neurons per unit time during training.
}
\label{fig:general-comparison}
\vspace{-0.4cm}
\end{figure*}

\subsection{Communication Load of {\MBSNN} and {\IWSNN}}

From the descriptions of {\MBSNN} and {\IWSNN} given above, they require bi-directional communication. For both rules, at each time $t$, unicast communication from neurons to CP is required to collect information $\{\{\ell\big(x_{i,t},\sigma(u_{i,t}^k)\big)\}_{k=1}^K\}_{i \in \set{X}}$ from all visible neurons. It entails unicast communication load $\CNtoCP = K|\set{X}|$ real numbers, as for {\GEMSNN}. For {\MBSNN}, the collected information is used to compute the per-sample learning signals $\{v_{\bmtheta^\text{X},t}^{k,\gamma}\}_{k=1}^K$, which are then sent back to all hidden neurons, resulting a broadcast communication load from CP to neurons equal to $\CCPtoN = K|\set{H}|$. In contrast, for {\IWSNN}, the collected information is used {\em (i)} to compute the importance weights $\{\bmsigma_{\text{SM}}^k\big( \bmv_{\bmtheta^\text{X},t}^{\gamma}\big)\}_{k=1}^K$ of the samples, which are sent back to all visible neurons; and {\em (ii)} to compute the scalar learning signal $\log R_{\bmtheta^\text{X},t}^{K,\gamma}$ that is fed back to all hidden neurons. The resulting broadcast communication load is $\CCPtoN = K|\set{X}|+|\set{H}|$. As compared to {\GEMSNN} whose broadcast load is $K(|\set{X}|+|\set{H}|)$, {\MBSNN} and {\IWSNN} have smaller broadcast communication load (see Table.~\ref{tab:comparison} for details).

\subsection{Additional Experiments on Classification Task}

We provide a comparison among the multi-sample training schemes, including the proposed {\GEMSNN} and two alternatives {\MBSNN} and {\IWSNN}. 
We trained an SNN with $K$ samples by using {\GEMSNN}, {\MBSNN} and {\IWSNN}, and tested the SNN with $K_I=K$ samples, where the corresponding estimated log-loss, accuracy, broadcast communication load and the number of spikes for hidden neurons during training are illustrated as a function of $K$ in Fig.~\ref{fig:general-comparison}. We recall that the multi-samples learning rules use learning signals for visible and hidden neurons in different ways. 
For visible neurons, {\MBSNN} does not differentiate among the contributions of different samples, while {\GEMSNN} and {\IWSNN} weigh them according to importance weights. For hidden neurons, {\IWSNN} uses a scalar learning signal, while both {\GEMSNN} and {\MBSNN} apply {\em per-sample} learning signals in different way -- {\GEMSNN} applies the importance weights of samples that are normalized by using SoftMax function, and {\MBSNN} uses the unnormalized values.

From Fig.~\ref{fig:general-comparison}, we first observe that {\GEMSNN} outperforms other methods for $K > 1$ in terms of test log-loss, accuracy, and number of spikes, while requiring the largest broadcast communication load $\CCPtoN$. Focusing on the impact of learning signals used for visible neurons, it is seen that using the importance weights in {\IWSNN} and {\GEMSNN} improves test performance. For this classification task, where the model consists of a small number of read-out visible neurons, the difference in performance among the proposed training schemes is largely due to their different use of learning signals for the hidden neurons. 
Specifically, in terms of the estimated test log-loss, per-sample importance weights used in {\GEMSNN} that are normalized using the SoftMax among the $K$ samples outperforms the global scalar learning signal in {\IWSNN} and the per-sample (unnormalized) learning signal in {\MBSNN}; while the use of multiple samples in {\GEMSNN} is seen to enhance the performance for large $K$. We note that other two schemes {\MBSNN} and {\IWSNN} can be applied with arbitrary variational posterior parameterized by learnable parameters, which further can be optimized during training. With our choice of causally conditioned distribution \eqref{eq:causally-cond}, it is not feasible. 
Finally, it is observed that the proposed schemes provide different trade-offs between costs (in terms of communication load and energy consumption) and learning performance (in terms of test log-loss and accuracy).




\bibliographystyle{IEEEtran}
\bibliography{ref}

\end{document}